\documentclass[letterpaper, 10 pt, journal]{IEEEtran}
\usepackage[utf8]{inputenc}

\usepackage{amssymb}
\usepackage{amsfonts}
\usepackage{amsmath}
\usepackage{cite}

\newcommand{\citep}[1]{\cite{#1}}
\newcommand{\citet}[1]{\cite{#1}}
\usepackage{scalerel}

\usepackage[table]{xcolor}
\usepackage[colorlinks=true,linkcolor=gray,citecolor=gray]{hyperref}
\usepackage{graphicx}

\usepackage{tabularx,calc} 
\usepackage{adjustbox} 
\usepackage{array}
\usepackage{multicol}
\usepackage{multirow,booktabs}

\newcommand{\R}{\mathbb{R}}

\newcommand{\traj}{(\cdot)}
\newcommand{\Env}{ {\rm Env} }

\newcommand{\qdot}{\nu}
\newcommand{\qddot}{\dot{\nu}}

\newcommand{\ddt}{\frac{\rm d}{ {\rm d} t} }

\DeclareMathOperator*{\minimize}{minimize}
\DeclareMathOperator{\st}{subject~to}

\newcommand{\com}{{\rm CoM}}
\newcommand{\cop}{{\rm CoP}}

\title{Optimization-Based Control for\\ Dynamic Legged Robots}
\author{
Patrick M.~Wensing$^1$,
Michael Posa$^2$,
Yue Hu$^3$,
Adrien Escande$^4$,
Nicolas Mansard$^5$,
Andrea Del Prete$^6$\\[-4ex] ~
\thanks{$^{1}$University of Notre Dame: 
        {\tt\footnotesize pwensing@nd.edu}}%
\thanks{$^{2}$University of Pennsylvania:
        {\tt\footnotesize posa@seas.upenn.edu}}%
\thanks{$^{3}$University of Waterloo:
        {\tt\footnotesize yue.hu@uwaterloo.ca}}%
\thanks{$^{4}$INRIA:
        {\tt\footnotesize adrien.escande@inria.fr}}%
\thanks{$^{5}$LAAS-CNRS:
        {\tt\footnotesize nmansard@laas.fr}}%
\thanks{$^{6}$University of Trento:
        {\tt\footnotesize andrea.delprete@unitn.it}}%
}

\begin{document}

\maketitle

\begin{abstract}
In a world designed for legs, quadrupeds, bipeds, and humanoids have the opportunity to impact emerging robotics applications from logistics, to agriculture, to home assistance. The goal of this survey is to cover the recent progress toward these applications that has been driven by model-based optimization for the real-time generation and control of movement. The majority of the research community has converged on the idea of generating locomotion control laws by solving an optimal control problem (OCP) in either a model-based or data-driven manner. However, solving the most general of these problems online remains intractable due to complexities from intermittent unidirectional contacts with the environment, and from the many degrees of freedom of legged robots. This survey covers methods that have been pursued to make these OCPs computationally tractable, with specific focus on how environmental contacts are treated, how the model can be simplified, and how these choices affect the numerical solution methods employed. The survey focuses on model-based optimization, covering its recent use in a stand alone fashion, and suggesting avenues for combination with learning-based formulations to further accelerate progress in this growing field.
\end{abstract}

\setcounter{section}{0}

\section{Introduction}
\label{sec:intro}
\nocite{Tassa2012,koenemann2015whole,Neunert2017a,neunert2016fast,Neunert2017,Chatzinikolaidis2021,Cleach2021, Howell2022, Onol2020}
\nocite{Zhu2021}
\nocite{Mordatch2012}
\nocite{Carius2018, Kong2021}
\nocite{Posa2014,xi2014optimal,xi2016selecting,Manchester2019a,Patel2019,Doshi2019,Aydinoglu2022}
\nocite{Wampler2009,Ibanez2014,Dai14,Ponton2016a,Winkler2018,Aceituno2018,Ding2018}
\nocite{Morimoto2003,Geisert2017,Budhiraja2019,Budhiraja2019a,Li2020hybrid}
\nocite{schultz2009modeling,Mombaur2009,koch2014optimization,hereid2015hybrid}
\nocite{Ratliff2009a,Posa2016,Hereid16,pardo2017hybrid}
\nocite{Carpentier2018,Budhiraja2019}
\nocite{Feng2013,farshidian2017efficient,Grandia2019a}
\nocite{Dai2012,Audren2014a,Dai2016,Caron2016,Kuindersma2016,Xiong2020,Khadiv2020,Fernbach2020}
\nocite{Kajita2003,Wieber2008,Herdt2010,Mordatch2010,Wieber14,Brasseur2015,Hopkins2015a,Liu2016b,Herzog2016a,DiCarlo2018a}

Over the past decade, we have witnessed rapid growth in the capabilities of legged robots, transitioning from a state of the art where only a few research groups had access to capable platforms, to one where robust locomotion is now common in industry and academic laboratories. This rapid progress has been enabled by a combination of advances across design and control, with optimization-based control strategies playing a central role in many of the milestone demonstrations during this period. Through these advances, current bipeds, quadrupeds, and humanoids can now walk reliably in nominal environments, and these improved capabilities have led to the first practical deployments of legged systems (e.g., the robot Spot from Boston Dynamics). In a world built for legs, the growing scope of these deployments offers a broad opportunity for impact on applications spanning logistics (e.g., delivery), agriculture, and home assistance, among many others. 

Across these applications, optimization-based control strategies offer many advantages as a pathway to (or component of) capable autonomy. Two prevailing types of optimization-based control have been pursued: predictive and reactive. Predictive controllers consider an explicit system model to ``reason'' (in a strictly mathematical sense) about the consequences of their actions, iteratively devising and improving motion plans in response to the situation at hand. Reactive controllers, by contrast, only consider their actions for the current instant. The execution of these controllers may depend on the system model, or may be model-free (e.g., the result of offline policy optimization). 
The benefits of the model-based approach are that it allows one to naturally consider safety constraints, which will be critical for many of the envisioned applications for legged robots. An additional motivation is that first-principles models generalize well for performance in unforeseen situations, which will be an asset for flexible autonomous deployments. The downsides of the model-based approach are that any safety certificates or performance guarantees are only as good as the models they are based on, and there is a fundamental challenge of incorporating perceptual streams into problem formulations. As such, while the past decade of model-based advances has accelerated the progress of the legged locomotion community, many recent results have motivated the benefits of learning-based strategies for gracefully handling hard-to-model and perception aspects. With this big picture in mind, a central goal of the review is to synthesize the past decade of model-based advances so that they can be further improved in the future and also incorporated with learning-based strategies---ultimately accelerating the practical deployment of legged robots. 

    
    

\begin{figure}[t]
\center
    \includegraphics[width=.8 \columnwidth]{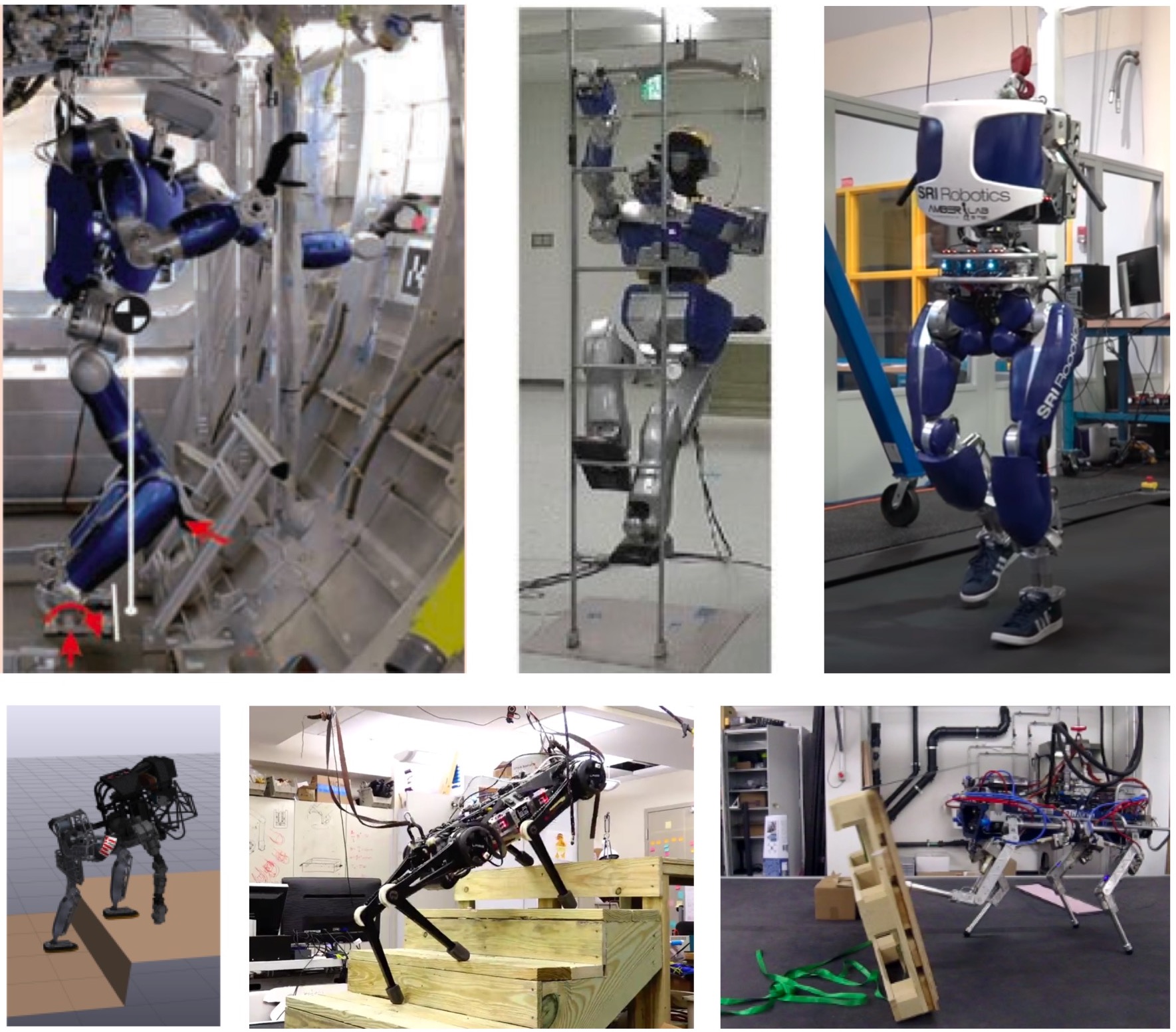}
    \caption{Optimization-based control has been a common enabler for legged systems to handle complex environments \cite{kheddar2019humanoid,Vaillant2016,Hereid16,Posa2014,bledt2018cheetah,Neunert2017a}.}
    \label{fig:OBCexamples}
    \vspace{-5px}
\end{figure}


\subsection{Historical Context for the Survey}

Much of the recent progress in optimization-based control for robots rests on decades of progress from the mathematical programming community and its early translation to our field. 
The idea of using optimization for the specification and control of robot movement traces back to decades before its current popularity (e.g., \cite{Slotine1985a}). The 1990s saw the use of quadratic programming for torque control of redundant manipulators \cite{Cheng1995}, with the 2000s seeing generalizations of these ideas for application to the control of humanoids in simulation \cite{Collette2007, Park2007,Abe2007}. 
The advent of polynomial-time solutions
to important classes of convex optimization problems in the mid-1990s \cite{Nesterov1994} and subsequent commercially available solvers (e.g., \cite{Andersen2003}) has contributed to the rise in maturity and accessibility of optimization solutions for control. 
Building on these advances, work in the late 2000s and in the wave of research before the 2015 DARPA Robotics Challenge (DRC) focused on convex optimization for maintaining balance via reactive control. The role of optimization in these accomplishments has since led to 
a growth of recent work for pushing the boundaries of predictive control for quadrupeds and humanoids executing broader dynamic locomotion in more challenging environments. 

\subsection{Survey Goals}

The overarching goal of this survey is to provide a synthesized entry point into this recent body of work. 
As such, our intention is for the survey to serve as the most effective resource for early-stage graduate students, while also providing new perspectives and targeted follow-on reading for established experts. 
Work since the DRC has led to many new methods for optimization through contacts, for real-time optimization using new simplified models, and through advances to numerical methods for handling ever more complicated robot models. These advances are the primary focus of the review. 

Beyond covering this past work, an additional motivation for this survey is to provide the background for the next steps in legged locomotion research, enabling these systems to move beyond nominal environments and unlock mobility in unstructured terrains. A great deal of current research interest focuses on how to combine machine learning strategies with previously developed model-based ones. In this sense, the survey will provide valuable background for machine learning practitioners to come up to speed on existing model-based approaches. Other research continues to push forward numerical methods and algorithms for legged robots to make informed model-based decisions. 
The survey should serve as a valuable overview of recent work for these groups as well.




\section{Problem Statement and Overview}
\label{sec:formulation}

During the last decade, most of the research community working on legged robots (humanoids and quadrupeds in particular) has converged to the idea of generating motion based on the formulation of an Optimal Control Problem (OCP). 
In a high-level form, such an OCP can be written as:
\begin{subequations}
\begin{align} 
    \minimize_{ x\traj , u\traj, \lambda\traj} 
            ~& {\rm Cost}( x\traj, u\traj, \lambda\traj) \label{eq:ocp_cost} \\[.15ex]
    \st     ~& \scaleobj{.92}{M(q) \, \qddot + C(q,\qdot) \, \qdot + \tau_g(q) = S^T \tau + J(q)^T \lambda} \label{eq:ocp_wb_dyn} \\[.15ex]
             & {\rm ContactConstraints}(x(t), \lambda(t), u(t), \Env)  \label{eq:ocp_contact_constr}  \\[.15ex]
             & {\rm KinematicsConstraints}(x(t)) \label{eq:ocp_kin_constr} \\[.15ex]
             & {\rm InputConstraints}(u(t), x(t) ) \label{eq:ocp_input_constr} \\[.15ex]
             & {\rm TaskConstraints}(x(t), u(t), \lambda(t) ) \label{eq:ocp_task_constr} \quad \quad \forall t\,.
\end{align}
\label{eq:ocp_wb}
\end{subequations}
The (infinite-dimensional) decision variables of the OCP are the {\em trajectories} of state $x=(q, \qdot)$, control $u\triangleq \tau$ and contact forces $\lambda$ exchanged between the robot and the environment Env. The robot configuration is represented as $q$, whereas its velocity as $\qdot$.
Eq.~\eqref{eq:ocp_cost} is a user-defined cost functional representing a metric to minimize, such as energy consumption or distance to a desired target.
Eq.~\eqref{eq:ocp_wb_dyn} represents the nonlinear whole-body robot dynamics~\citep{featherstone2008rigid}, where $M(q)$ is the mass matrix, $C(q, \qdot) \qdot$ accounts for Coriolis and centrifugal forces, $\tau_q(g)$ contains the gravity forces, $S$ is typically a matrix that selects the actuated degrees of freedom, and $J(q)$ is the contact Jacobian.
Eq.~\eqref{eq:ocp_contact_constr} encodes the contact-related constraints, such as non-penetration of rigid objects and friction-cone force constraints.
Eq.~\eqref{eq:ocp_kin_constr} includes state constraints resulting from the kinematics, such as joint position and velocity limits.
Eq.~\eqref{eq:ocp_input_constr} typically includes the motor torque limits, but it could in general represent any input constraint.
Finally, \eqref{eq:ocp_task_constr} can be used to include any task-specific constraint in the OCP, such  as fixed initial/final state, or field-of-view limits for a camera.

Clearly, this OCP is very general and therefore allows one to encode a wide variety of movements.
For instance, point-to-point locomotion can be generated by specifying the initial and final robot states~\citep{Mombaur2009}.
An object manipulation behavior could instead be computed by including the state of the object in $x$ and specifying its desired value with an appropriate cost function~\citep{Ratliff2009a, Toussaint2018, Hogan2017}.
While appreciating its versatility, we should acknowledge that some intrinsic limitations do exist in the OCP approach. 
For instance, it is hard to ensure stability~\citep{Mombaur2009} or robustness for the computed trajectories \citep{Dai2012}.
However, this review focuses on another limitation and how to address it: the computational complexity of the problem.
Indeed, problem~\eqref{eq:ocp_wb} hides several challenges, the main ones being: non-smoothness/stiffness of the dynamics, nonconvexity, and dimensionality. 
The rest of this section acts as an executive summary of the rest of the paper by briefly discussing different approaches to tackle these challenges and make problem~\eqref{eq:ocp_wb} computationally tractable  for offline trajectory optimization, or for online model predictive control (MPC). Each aspect is analyzed more thoroughly in Sections~\ref{sec:contacts}-\ref{sec:control}.

\subsection{Contact Models}
The first issue (non-smoothness/stiffness), which is discussed in detail in Section~\ref{sec:contacts}, arises from physical contacts modeled in \eqref{eq:ocp_contact_constr}.
Contacts can be modeled as either \emph{rigid} or \emph{visco-elastic}.
With appropriate choices, visco-elastic models ensure continuous dynamics, and smoothing techniques can be used to make them  differentiable. 
The downside of visco-elastic models is that large stiffness values are necessary to generate realistic behaviors. 
This feature leads to stiff differential equations with corresponding numerical challenges for simulation and optimization.

Alternatively, contacts can be modeled as rigid (i.e., no penetration is allowed).
In this case, the dynamics becomes \emph{hybrid} because when two points make contact, their relative velocity must immediately become zero to avoid penetration.
Therefore, the robot dynamics must be described by a mix of continuous-time and discrete-time equations, i.e., as a hybrid dynamic system.
The resulting OCP can then be tackled either as a Linear Complementarity Program (LCP)~\citep{Stewart1996a,Posa2014,Patel2019} or as a Mixed Integer Program (MIP)~\citep{Aceituno2018,Deits14}, both of which require customized optimization techniques that are typically much less efficient than classic smooth optimization strategies. 

A common way around the non-smoothness introduced by rigid contacts is to let the user fix the order in which contacts are made and broken.
This makes the system dynamics time-varying (a special case of hybrid dynamics) and the OCP differentiable, therefore efficient smooth optimization can be used.
The obvious downside is that it may be hard to guess the contact phases.

\subsection{Dynamic Models}
Besides the contact dynamics, the other major source of complexity is the robot multi-body dynamics~\eqref{eq:ocp_wb_dyn}, which is high-dimensional and nonlinear.
In turn, this makes the resulting optimization problem high-dimensional and non-convex.
The high dimension is especially concerning in the context of online optimization, where fast computation times are mandatory.
The non-convexity instead is always concerning, as it makes the solver sensitive to the initial guess that is used. 
As partial or total remedy to these issues, several \emph{simplified models} have been proposed in the literature (also known as reduced-order models, or template models), and they are discussed in Section~\ref{sec:simplified_models}.
These models should capture the most important part of the robot dynamics with a reduced state size.
For example, for locomotion, the widely-used Linear Inverted Pendulum (LIP) model~\citep{Kajita2003} considers only the Center of Mass (CoM) of the robot and the contact locations, neglecting the details of joint angles and velocities.
Other models capture more features of the robot state, at the price of higher state size and nonlinearity in the dynamics.
While these simplified models are key enablers for fast online computation, their simplifying assumptions (e.g., constant CoM height), or neglected constraints~\citep{Carpentier2017} (e.g., joint position and torque bounds) can severely limit the generated motions.

\subsection{Optimal Control Solution Methods}
After the OCP is mathematically formulated, there are a range of design choices that remain for solving it. General optimal control methods can be categorized as direct methods, indirect methods, or those based on dynamic programming. Within robotics, direct optimization methods prevail almost universally. However, even after choosing to use a direct method, the optimization formulation needs to be approximated as a finite-size nonlinear program to be solved by a numerical solver. This approximation process is called {\em transcription} and it crucially affects the results in terms of accuracy, numerical stability, and computational complexity.  

Two families of transcription methods have been used in robotics (shooting and collocation) and we discuss them in Section~\ref{sec:transcription}. A shooting method known as Differential Dynamic Programming (DDP) has been a topic of recent interest in the community due to its favorable computational properties and provision of a locally optimal feedback policy in addition to a locally optimal open-loop control. More generally, but specific to direct optimization for locomotion, we examine how contact modeling choices affect the choice of transcription tools and algorithms available, which becomes most critical when optimization is left to freely choose contact sequences.

\begin{figure*}[t]
    \centering
    \includegraphics[width=.56 \textwidth]{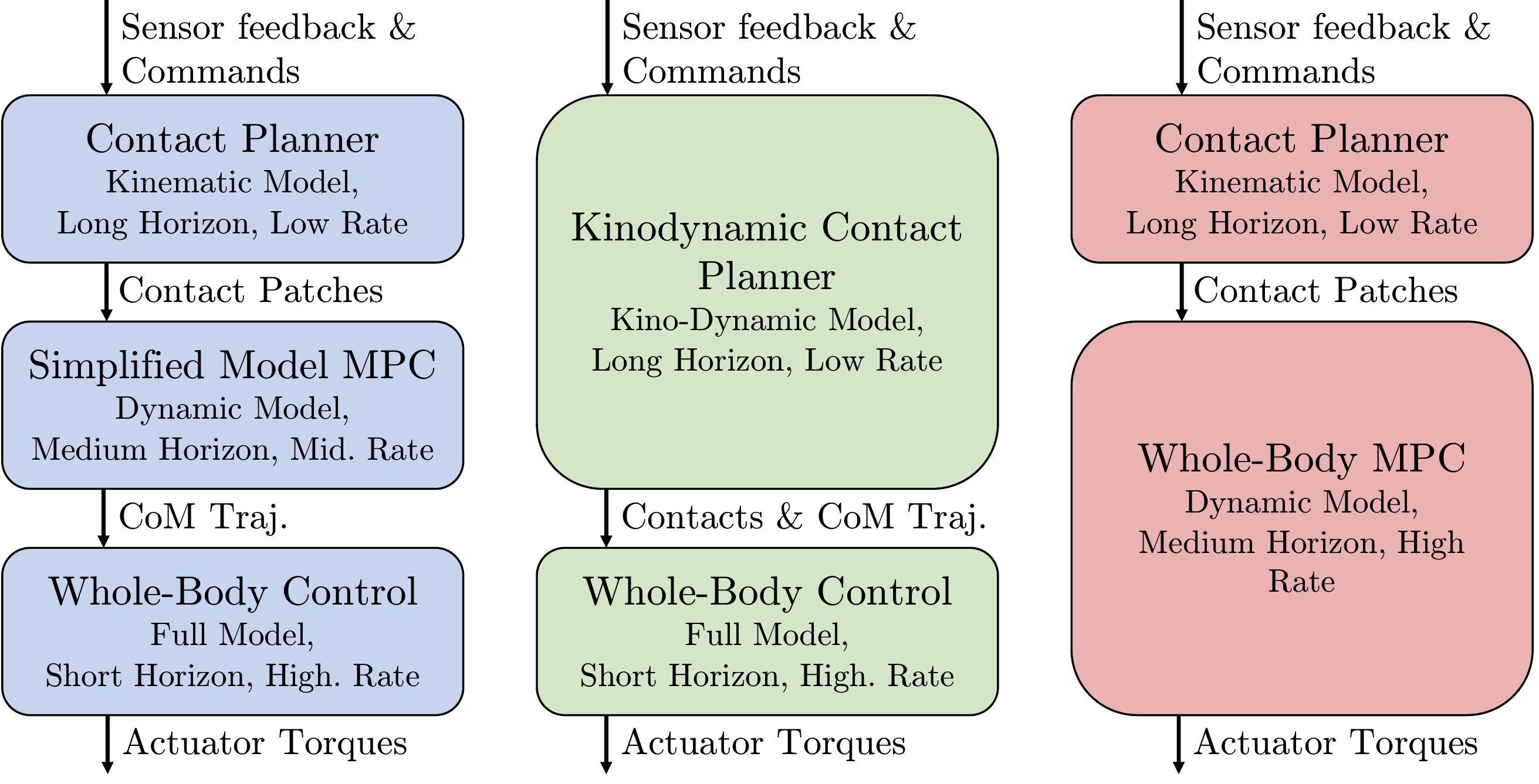}
    \caption{Different potential strategies for breaking Problem \ref{eq:ocp_wb} into smaller pieces to ease computational challenges.}
    \label{fig:breakdown}
    \vspace{-5px}
\end{figure*}

Table \ref{tab:SOTA2} serves to categorize the choices that state-of-the-art papers have used in adopting different approaches for modeling the contacts and their sequencing, for simplifying the dynamics, and, ultimately, for numerically solving variants of \eqref{eq:ocp_wb}. In practice, the highlighted approaches play a key role in control designers breaking problem~\eqref{eq:ocp_wb} into many smaller subproblems that focus on building an overall solution in a hierarchical fashion (e.g., solving for footsteps first, then optimizing motions with fixed footsteps). Methods for architecting this decomposition remain an art more than a science, with a few potential approaches depicted in Fig.~\ref{fig:breakdown}.

\begin{table}[!tbp]
\caption{Overview of the state of the art.}
\centering 
\scaleobj{.869}{
\begin{tabularx}{1.15\columnwidth}{X | m{1.5cm} m{1.5cm} m{1.5cm} m{2.1cm}}
\hline 
Papers       & Contact model & Gait & Dynamics & Transcription 
\\ \rowcolor[gray]{.9} 
\hline
\cite{Tassa2012,koenemann2015whole,Neunert2017a,neunert2016fast,Neunert2017,Chatzinikolaidis2021,Cleach2021, Howell2022, Onol2020}
& Soft/Smoothed & Optimized & Full & DDP 
\\
\cite{Zhu2021}
& Soft/Smoothed & Optimized & Full & Collocation \\
\cite{Mordatch2012}
& Soft/Smoothed & Optimized & Simplified & Collocation
\\ \rowcolor[gray]{.9} 
\cite{Carius2018, Kong2021}
& Rigid & Optimized & Full & DDP 
\\
\cite{Posa2014,xi2014optimal,xi2016selecting,Manchester2019a,Patel2019,Doshi2019,Aydinoglu2022}
& Rigid & Optimized & Full & Collocation
\\ \rowcolor[gray]{.9} 
\cite{Wampler2009,Ibanez2014,Dai14,Ponton2016a,Winkler2018,Aceituno2018,Ding2018}
& Rigid & Optimized & Simplified & Collocation
\\ 
\cite{Morimoto2003,Geisert2017,Budhiraja2019,Budhiraja2019a,Li2020hybrid}
& Rigid & Fixed & Full & DDP 
\\ \rowcolor[gray]{.9} 
\cite{schultz2009modeling,Mombaur2009,koch2014optimization,hereid2015hybrid}
& Rigid & Fixed & Full & Multiple Shooting
\\
\cite{Ratliff2009a,Posa2016,Hereid16,pardo2017hybrid}
& Rigid & Fixed & Full & Collocation
\\ \rowcolor[gray]{.9} 
\cite{Carpentier2018,Budhiraja2019}
& Rigid & Fixed & Simplified & Multiple Shooting 
\\ 
\cite{Feng2013,farshidian2017efficient,Grandia2019a}
& Rigid & Fixed & Simplified & DDP
\\ \rowcolor[gray]{.9} 
\cite{Dai2012,Audren2014a,Dai2016,Caron2016,Kuindersma2016,Xiong2020,Khadiv2020,Fernbach2020}
& Rigid & Fixed & Simplified & Collocation
\\ 
\cite{Kajita2003,Wieber2008,Herdt2010,Mordatch2010,Wieber14,Brasseur2015,Hopkins2015a,Liu2016b,Herzog2016a,DiCarlo2018a}
& Rigid & Fixed & Simplified & Single Shooting
\\
[0.5ex] \hline 
\end{tabularx}} \label{tab:SOTA2} 
\end{table}

\subsection{Trajectory Stabilization via Whole-Body Control}
Depending on the complexity of the formulated OCP, the computation time may be too large for it to be solved inside a fast control loop in MPC style. 
In these cases, a reactive stabilizing controller is required to execute the computed motion on real hardware, or to provide control of aspects of the system that were ignored during trajectory optimization (e.g., due to modeling simplifications, as with the first two configurations in Fig.~\ref{fig:breakdown}). 
Many techniques can be used to track a reference motion on a robotic system. 
However, in the last decade, the legged robotics community has converged to a certain class of reactive optimization-based whole-body control techniques, which mainly rely on the fast solution of small convex Quadratic Programs (QPs) to compute motor commands as a function of state feedback. 

These QP techniques have represented an evolution of previous operational-space control paradigms that were originally developed for the control of manipulators. However, for legged systems, new problems related to contact constraints, impacts, self-collisions, etc., have motivated a broader perspective on this classical problem. This evolution has led to the consideration of tasks that include fully-specified task-space motions, as well as constraints characterized by task-space inequalities. In spite of these many extensions, the formulation of reactive control as an instantaneous control strategy has enabled the method to retain a convex formulation that is lost in more general transcription strategies. The convex-optimization perspective on reactive control has also enabled other control paradigms, conventionally separate from operational-space control, to be incorporated as well. We describe these techniques and their connections in Section~\ref{sec:control}.

\section{Contact}
\label{sec:contacts}

 Computationally tractable treatment of physical, frictional contact with the environment, which leads to stiff and/or discontinuous equations of motion, is a fundamental challenge in optimization-based control for robotics and a primary distinction between \eqref{eq:ocp_wb} and standard optimal control problems.
 As a result, design decisions for (i) how to model the effects of contact and (ii)  the sequencing or scheduling of contact events play a significant role in distinguishing different approaches to solving \eqref{eq:ocp_wb}.
 Specifically, the details in this section center on how the choice of contact model affects parameterizations of the force $\lambda$ and the form of the contact constraints in \eqref{eq:ocp_contact_constr}.
 
This section is comprised of two parts. In Section~\ref{sec:modeling_contact}, we outline common techniques, and associated numerical challenges, for numerical models of contact dynamics. In Section~\ref{sec:contact_schedule}, we describe how the need to sequence contact events affects algorithmic choices and the modeling choices of Section~\ref{sec:modeling_contact}.

\subsection{Modeling contact}
\label{sec:modeling_contact}
Highly accurate mechanics models define contact as \textit{visco-elastic}, representing local deformations of surfaces and the corresponding stress-strain relationships.
While some approaches in robotics, particularly in soft robotics (e.g., \cite{Duriez2013}), do attempt to accurately represent deformation, where this deformation prevents interpenetration between bodies, rigid-body approximations to contact are most commonly used and will be the focus of this review. With this perspective, while bodies cannot deform, common visco-elastic contact models permit some penetration between contacting bodies, and the resulting contact forces are modeled as a function of penetration depth, relative normal and tangential velocities, and material properties.
Noting critical definitions here, we write $\phi(q)$ to be the vector-valued signed distance function, for all possible contacts. $J_n(q)$ and $J_t(q)$ are the normal and tangential Jacobians, such that $J_n(q)\nu$ and $J_t(q)\nu$ are the contact frame velocities.
Normal and tangential forces are similarly decomposed into $\lambda_n$ and $\lambda_t$, and, where convenient, we use $J(q)$ and $\lambda$ to represent  stacked Jacobians and forces.
The contact force, therefore, can be expressed as some function
\[
\lambda = F_{\rm contact}(\phi, J\nu).
\]
Common methods treat this function as a nonlinear spring and damper (e.g., \cite{Marhefka1999a, Drake2016}), where unilateral constraints determine when the spring releases; typically, contact forces cannot ``pull'', constraining $\lambda_n \geq 0$ and so separation occurs if $\lambda_n < 0$.
However, to accurately represent the interaction between rigid bodies, $F_{\rm contact}$ is numerically stiff~\cite{Ascher1998}, inheriting the mechanical stiffness of the material properties in the robot.
When making and breaking contact ($\phi = 0$), $F_{\rm contact}$ may also be non-differentiable. 
Frictional effects exhibit similar issues: in dry Coulomb friction, $\lambda_t$ depends on the sign of the tangential velocity, which is discontinuous.

\begin{figure}
    \centering
    \includegraphics[width=.82\hsize]{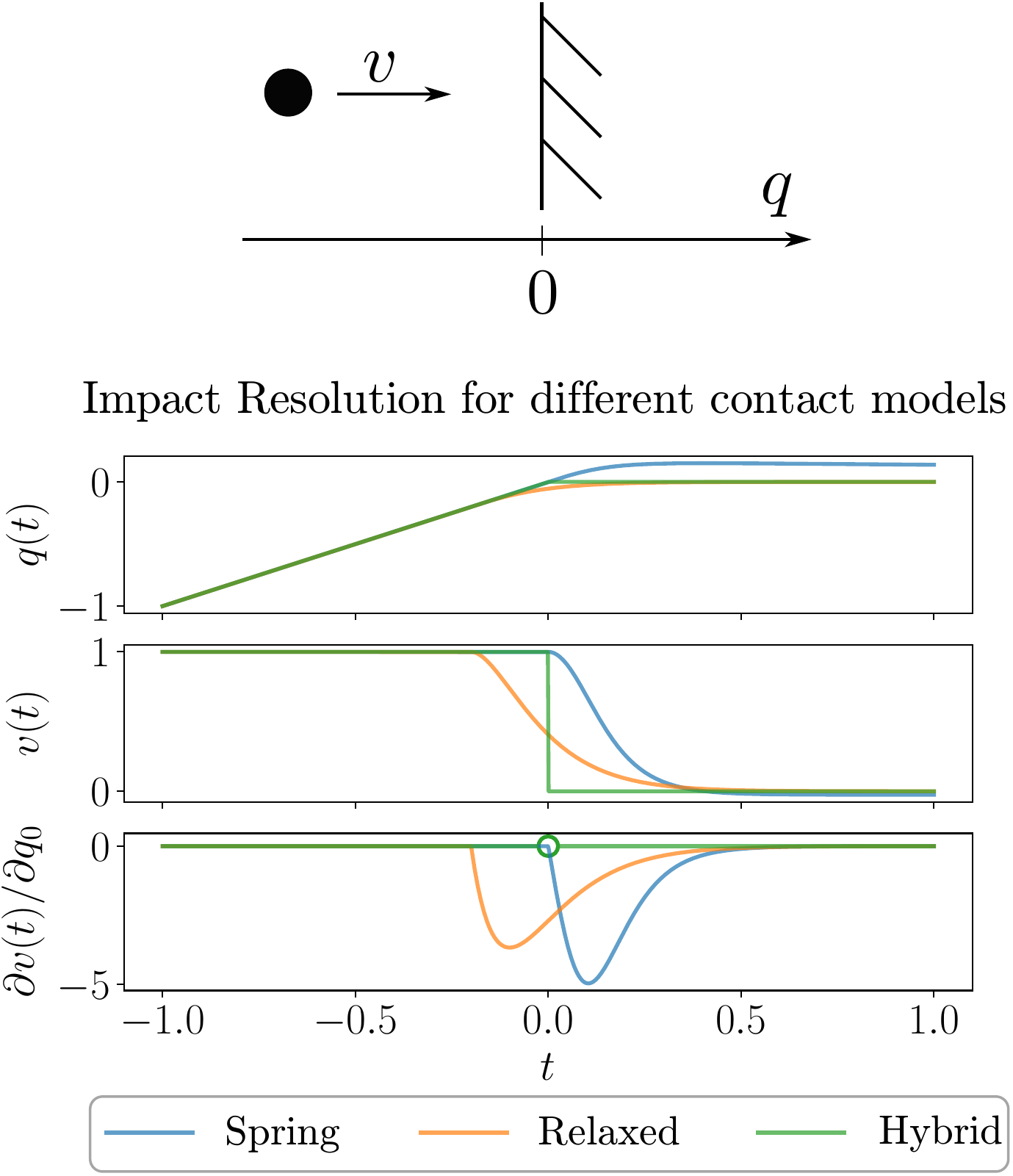}
    \caption{In an illustrative example, a particle heads toward a wall, with inelastic impact when $q=0$. Three common contact models are compared, showing (top-to-bottom) position, velocity, and sensitivity of velocity to initial conditions. (Blue) A Hunt-Crossley spring model \cite{Hunt1975} activates when contact occurs at $q(t)=0$, allowing some penetration. The sensitivity ${\partial v(t)}/{\partial q_0}$ is well-defined, though not smooth ($C^\infty$) at the impact. (Orange) a relaxed (softened) complementarity model, like MuJoCo, is smooth, with contact forces occurring prior to contact \cite{Todorov2014}. (Green) a hybrid model with inelastic impact captures rigid behavior with an instantaneous jump in velocity, noting that ${\partial v(t)}/{\partial q_0}$ is undefined at impact.
    All models have well-defined gradients \textit{away} from the impact event, enabling effective differentiation in these regimes. For the hybrid and spring model, these gradients are zero prior to contact; for the relaxed model, as it is smooth, these gradients are non-zero everywhere.}
    \label{fig:contact_models}
\end{figure}

Simply introducing these stiff or discontinuous dynamics to \eqref{eq:ocp_wb} leads to a poorly conditioned optimization problem and is generally avoided.
Algorithmic approaches, therefore, tend to focus on one of two choices: (i) \textit{infinitely} stiff representations, as hybrid models, or (ii) smoothing or softening  $F_{\rm contact}$ to improve numerical performance.
An illustrative example of a single contact is shown in Fig.~\ref{fig:contact_models}, demonstrating the resulting state trajectory and, critical for optimization-based control, the sensitivity of the solution with respect to initial conditions.

\subsubsection{Hybrid dynamics}
In the limit of infinite stiffness, when contact is initiated, the resulting forces become \textit{impulsive}; impacts, therefore, cause an instantaneous jump in velocity.
While a full description of hybrid systems is outside the scope of this review (see \cite{VanDerSchaft2000} for an overview), we briefly note the principles here.
Hybrid systems are defined by \textit{modes}, \textit{guards}, and \textit{resets}.
In this context, the mode is the contact state: identifying which objects are touching, and whether they are
sticking or sliding.
Guards determine when mode transitions occur, for instance making or breaking contact, and the reset map $R$ specifies the result of the transition (e.g., impact):
\[
x^+ = R(x^-).
\]
Within a given mode, the dynamics are well-defined and differentiable; thus, this formulation compresses the complexity of contact to the guards and transition events.
Referring to the root optimal control problem \eqref{eq:ocp_wb}, a hybrid formulation typically explicitly splits the decision variables $x(\cdot), u(\cdot), \lambda(\cdot)$ by mode, introducing boundary value constraints at the resets when numerically transcribing the dynamics in Section~\ref{sec:transcription}.


\subsubsection{Complementarity models}
A mathematically equivalent formulation of hybrid contact dynamics relies upon \textit{complementarity constraints} \cite{Stewart2000, Brogliato1999}, where a relationship between $\lambda$ and $x$ implicitly defines the hybrid dynamics.
For example, a non-penetration condition $\phi(q) \geq 0$, combined with $\lambda_n \geq 0$, and the property that forces can only be non-zero when in contact leads to the complementarity constraint
\[
0 \leq \lambda_n \perp \phi(q) \geq 0\,,
\]
where $\lambda_n \perp \phi(q)$ means that the vectors $\lambda_n$ and $\phi(q)$ must be orthogonal, which is equivalent to $\lambda_n^T \phi(q) = 0$.
This formulation can be expressed in time-stepping \cite{Stewart1996a} or continuous \cite{Anitescu97} models, with a similar set of complementarity constraints used to capture Coulomb friction.


\subsubsection{Pathologies}
Hybrid formulations of multi-contact robotics, while effective for simulation and control, can exhibit certain pathologies where solutions do not exist or where infinite solutions are possible \cite{Stewart2000, Remy2017, Halm2019}.
These challenges are the limiting case of the high sensitivity to initial conditions seen in the stiff differential equations.
From the perspective of optimization-based control, this stiffness may be directly represented in the stiffness of the underlying optimization problem.
Alternatively, for instance when using hybrid or complementarity models, planning algorithms that assume uniqueness therefore implicitly (perhaps even unbeknownst to the algorithm designer) select from the set of possible solutions (a property, for instance, of \cite{Posa2014}).
In many scenarios, this selection is benign; in others, it indicates that the resulting trajectory is practically impossible to track, due to the non-uniqueness (equivalently, stiffness).
Non-uniqueness can occur in a number of scenarios; most commonly, if $J(q)$ is rank deficient. Think, for example, of the static indeterminacy of a four-legged table or a flat robot foot on the ground, where the normal pressure distribution is indeterminate.

\subsubsection{Hybrid differentiability}
\label{sec:hybrid_differentiability}
In the hybrid formulation, while both the resulting trajectories $x(t)$ and the equations of motion may be discontinuous, under certain circumstances it is possible to generate well-conditioned derivatives of $x(t)$ with respect to initial conditions and control actions
\cite{Saccon2014, Burden2016} (see Fig.~\ref{fig:contact_models} for an example).
Typically, these derivatives exist if the sequence of modes is constant; the event times themselves might change, but the order does not and all transitions are inevitable.
When deviations might change the mode sequence, resulting trajectories are often \textit{not} differentiable with respect to initial conditions or parameterizations.
For example, if a trajectory $x(t)$ does not make contact, then no amount of local differentiation provides insight into the effects of initiating contact; put more explicitly, solutions $x(t)$ are non-analytic in their initial conditions and control inputs. This property poses natural challenges for optimizers to discover new contacts through local information alone, and this motivates many relaxations or other tailored strategies for scheduling contacts.

To mitigate this issue, certain simulators ensure global differentiability or even smoothness (termed \textit{differentiable simulation}, e.g., MuJoCo \cite{Todorov2014} or TDS \cite{Heiden2021}).
While these methods do consistently generate local gradients, this inevitably ties the accuracy (stiffness) of the underlying model to stiffness in the resulting optimization problem.
The precise nature of this trade-off and the relevance of the different modeling inaccuracies (see, e.g., \cite{Erez2015, Fazeli2017, Acosta2022}) remain unknown.

\subsection{Scheduling contact}
\label{sec:contact_schedule}
As the discussion of differentiability above implies, there are clear distinctions in computational tractability between versions of \eqref{eq:ocp_wb} with a fixed, known mode sequence and variations where the optimization problem must also determine the ordering.
In many robotics problems, this sequence may be clear; for example, bipedal walking over flat terrain typically follows a ``left foot, right foot'' ordering with minimal deviation.
In others, for instance, movements using whole-body contact or locomotion over varied surfaces with multiple potential foothold locations, the challenge of finding an optimal ordering may dominate the control problem.

\subsubsection{Known modes sequences}
In this setting, once discretized (see Section~\ref{sec:transcription}), the hybrid optimal control problem is differentiable with respect to the transcribed decision variables.
Within this framework, distinctions exist between methods in representation of the hybrid state space and the contact forces.
In some settings, particularly when deploying simplified models, it is common to use \textit{minimal coordinates} to represent the configuration of a robot in contact (e.g., \cite{Byl2008,Bhounsule2015}).
Contact points are transformed into pin joints, removing degrees of freedom and simplifying the resulting optimal control problem.
When contact modes change, however, the hybrid jump must also capture the change in dimension of the state space.
The use of minimal coordinates is computationally efficient, although it is difficult to impose constraints (e.g., friction) on $\lambda$ as inverse dynamics are necessary to reconstruct the constraint forces.
Furthermore, in multi-contact settings, where the contact points generate a closed kinematic chain, global definitions of minimal coordinates may not exist. 

The more general formulation utilizes the same maximal, or \textit{floating-base coordinates} for all contact modes.
The constrained dynamics then enforce that active contacts remain touching \cite{Mombaur2009, Budhiraja2019a}.
Some approaches directly compute the force, e.g., $\lambda(x,u)$ \citep{Budhiraja2019a}.
This approach has the advantage of relative simplicity, though heuristics are necessary when the rigid-body force is non-unique.
Alternate methods include $\lambda(t)$ as a decision variable (see \eqref{eq:ocp_wb}), along with corresponding constraints to ensure physical accuracy \cite{Posa2014, Manchester2019a}.
While more variables are needed in this lifted formulation, it can be better conditioned numerically.
It should also be noted, however, that imposing constraints \textit{differentially} in this fashion comes at the risk of constraint drift due to integration error.
Methods in this class can employ implicit integration schemes to eliminate this drift at some computational cost \citep{Posa2016, bordalba2022direct}.


\subsubsection{Hybrid sequence optimization}
\label{sec:contact:hybrid_seq_optimization}
As an immediate extension of the known-sequence scenario, one could jointly optimize over the discrete states of the hybrid sequence and the corresponding robot motion.
This can naturally be expressed via \textit{mixed-integer optimization} \cite{Ibanez2014, Aceituno2018}, or as a bilevel optimization problem \cite{Wampler2009}.
Mixed-integer programs (MIPs) have been used to compute contact sequences accounting for obstacle avoidance and step-to-step reachability \cite{Deits14} and approximated as computationally efficient L1-norm minimization (SL1M)~\cite{Song2021}. Other strategies ensure ``quasi-static'' feasibility by limiting the search to ``quasi-flat'' contact surfaces (i.e., surfaces where the friction cone contains the gravity direction) \cite{Tonneau2020}.
In the worst case, these methods must explore every possible mode sequence, and so are most effective when there are relatively few potential sequences or when effective heuristics are available to guide the search.

\emph{Sampling-based} techniques are a valid alternative to MIPs. For instance, rapidly-exploring random trees (RRTs) have been used to plan footsteps on flat ground, avoiding obstacles \cite{Perrin2012a}.
Similarly, probabilistic road maps (PRMs) have been used to plan a collision-free path for the robot's base, keeping it sufficiently close to the environment to allow for contact creation, while coupled with corresponding whole-body planning \cite{Tonneau2018a}. 

\begin{figure*}
\center
\includegraphics[width=.8\textwidth]{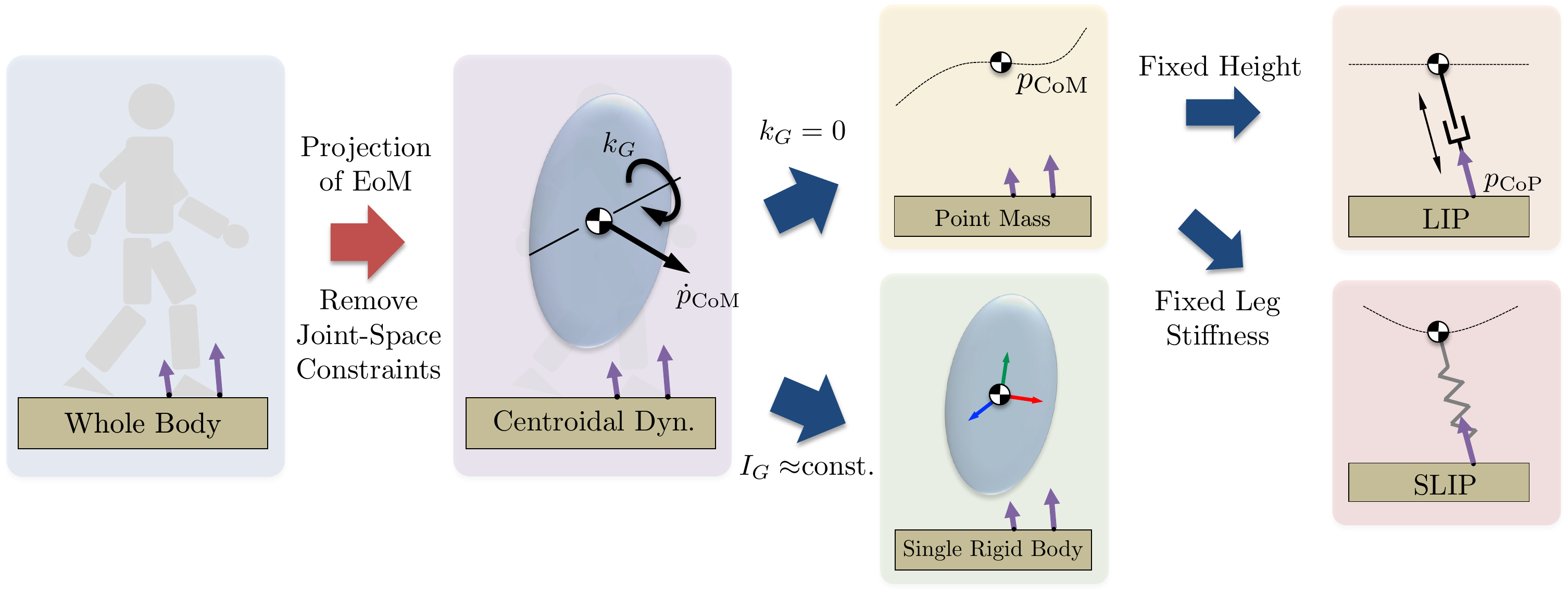}
\caption{The centroidal dynamics is a common modeling simplification that focuses on the CoM and net momentum. It is a special modeling simplification in that it exactly projects the whole-body dynamics, while relaxing joint-space constraints. Restrictions placed on this model then lead to a tree of other models. }
\label{fig:simple_models}
\end{figure*}

\subsubsection{Contact-implicit planning}
Contact-implicit (alternatively, contact-invariant) methods are variations on  hybrid optimization that embed the relationship between state and force into a nonlinear program, implicitly representing the hybrid mode without discrete variables.
Approaches here are most often based on either the complementarity formulation \cite{Posa2014} or on smooth approximations of the contact dynamics \cite{Mordatch2012,neunert2016fast}.
Prior to convergence, these methods typically violate strict complementarity, thus simultaneously exploring multiple contact sequences.
Contact-implicit optimization, however, can suffer from poor numerical conditioning and can require high-quality initial guesses.
Contact-implicit methods have found applications in a number of robotic domains, including model predictive control \cite{Tassa2012} and gait optimization for microrobots \cite{Doshi2019}.
Recent work in this area has focused on improving the numerical performance, for example by 
using simplified models \cite{Dai14,Winkler2018} or improving the accuracy of the numerical integration schemes \cite{Manchester2019a,Patel2019}.
While these methods are typically limited to offline computation of motion plans, progress has been made in real-time scheduling of contact \cite{Cleach2021, Aydinoglu2022}

These previous contact-implicit approaches explicitly account for the hybrid nature of planning through contact.
Alternatively, gradient-based methods, like iLQR or DDP (see Section~\ref{sec:ddp}), can be applied.
This is most common when coupled with a differentiable contact model, enabling the optimization method to ``discover'' new modes (when it happens to bump into them, perhaps guided by the smoothing in the differentiable model) \cite{Tassa2012, Onol2020, Chatzinikolaidis2021}.
Other methods maintain a rigid model, and consider gradient analyses around the contact sequence found during forward simulation (e.g., \cite{ Carius2018, Kong2021}), leveraging the almost everywhere differentiability discussed above.
Broadly speaking, these methods are typically highly sensitive to their initializations and struggle to discover contact sequences that vary dramatically from that of the initial guess.

\subsection{Summary}
When the sequence of environmental contacts can be known \textit{a priori}, then the choice of contact model should focus purely on physical realism, as nearly any transcription method can be applied without disruption.
It is substantially more difficult, however, to simultaneously plan motion and contact schedule.
While significant progress has been made over the last decade, contact planning remains the main challenge for generating arbitrary locomotion behaviors in complex environments. 
Ultimately, this challenge arises from our reliance on gradient-based optimization, which despite being the workhorse behind the progress discussed in this survey, is fundamentally unsuited for non-smooth contact-implicit problems.

\subsubsection{Relationship to learning} While this review focuses on model-based optimization, there is an implicit requirement that such a model must first be identified or learned.
Methods for learning and identification of contact align closely with the modeling choices detailed in this section.
When contact modes are known or can be easily identified, hybrid approaches are commonly used \cite{Fazeli2017}. Otherwise, machine learning can be directly applied to a smoothed or differentiable contact model (e.g., \cite{Belbute-Peres2018}). Alternatively, contact-implicit methods have led to data-efficient mechanisms for contact model learning \cite{Pfrommer2020}. 
Aside from identifying \textit{models}, machine learning has also shown promise in identifying (or providing high-quality guesses for) potential mode sequences for control \cite{deits2019lvis, cauligi2021coco}, where an effective pre-solve for the mode schedule substantially reduces the computational difficulty of the subsequent trajectory optimization.
We note, as well, that \textit{model-free} reinforcement learning (RL), which has been widely applied to similar problems in control (e.g., legged locomotion \cite{Siekmann2021, Miki2022}), typically also leverages smoothed (or stochastic \cite{Suh2022}) contact models for sim-to-real transfer.


\section{Simplified Models}
\label{sec:simplified_models}

A second source of complexity for the efficient solution of \eqref{eq:ocp_wb} arises from the whole-body dynamics model \eqref{eq:ocp_wb_dyn}. As noted earlier, the dynamics of legged systems are high-dimensional and nonlinear, and these features correspondingly make the optimization problem high-dimensional and nonconvex. These challenges motivate simplified models in place of \eqref{eq:ocp_wb_dyn} that capture its most salient features in a reduced set of differential equations. A central question is then how to select such a simplified model.  We use the evolution in Fig.~\ref{fig:simple_models} to walk through the most common modeling simplifications adopted.

The role that contacts play within the whole-body dynamics \eqref{eq:ocp_wb_dyn} has motivated the majority of simplified models employed to date. For a floating-base system, it is common to partition the generalized velocity as $\nu = (\nu_b, ~\nu_j)$ where $\nu_b=(\omega_b,~v_b)\in \R^6$ gives the angular and linear velocity of the floating base and $\nu_j$ the generalized velocity of the joints. 
With this partitioning in $\nu$, the dynamics are partitioned as
\begin{equation}
\begin{bmatrix} M_{bb} & M_{bj}\\ M_{jb} & M_{jj} \end{bmatrix} \begin{bmatrix} \dot{\nu_b} \\ \dot{\nu}_j \end{bmatrix} + C(q,\nu) \nu + \tau_g(q) = \begin{bmatrix} 0 \\ \tau \end{bmatrix} + J(q)^T \lambda\,.
\label{eq:partitioned_dynamics}
\end{equation}
Suppose now that the robot is given a desired motion $(q_d(t), \nu_d(t))$. Considering \eqref{eq:partitioned_dynamics}, the left-hand side of the equation is fixed along the desired motion, and so the motion is possible only when the equation can be satisfied through proper choice of joint torques $\tau$ and contact forces $\lambda$. If the robot has sufficiently strong actuators, then $\tau$ can be chosen to satisfy the bottom set of equations. However, since joint torques do not affect the first six rows, the main dynamic limitations on movement are determined by the contact force constraints (i.e., friction and unilaterality). 

\subsection{Centroidal Dynamics - Modeling and Optimization}
\label{sec:simplified_models:centroidal}

It can be shown that the first six rows of \eqref{eq:partitioned_dynamics} describe the evolution of the net linear and angular momentum of the system as a whole. This property follows from first principles of mechanics, and has been detailed within the context of floating-base robotics models in \citep{Wieber2006a,Wensing2016}. 
It is common to consider the angular momentum about the CoM denoted $k_G \in \R^3$, as this quantity 
is both conserved during free flight and empirically remains close to zero during human walking. Denoting the total linear momentum as $l_G \in \R^3$, these momenta compose the {\em centroidal momentum} $h_G = (k_G,~l_G) \in \R^6$ \citep{OrinGoswami13}. This quantity is related to the generalized velocities via
\begin{equation}
h_G = A_G(q)\, \nu\,,
\label{eq:CMM}
\end{equation}
where $A_G(q)$ is called the centroidal momentum matrix~\cite{OrinGoswami13}. Consider a case with $n_c$ contact points at locations $\{p_i\}_{i=1}^{n_c}$. It can be shown that the top rows of \eqref{eq:partitioned_dynamics} are equivalent to:
\begin{equation}
\dot{h}_G = \begin{bmatrix} \dot{k}_G \\ \dot{l}_G \end{bmatrix} = \begin{bmatrix} 0 \\ -M a_g \end{bmatrix} + \sum_{i=1}^{n_c} \begin{bmatrix} (p_i - p_{\com}) \times \lambda_i \\ \lambda_i \end{bmatrix}\,,
\label{eq:centroidal_dynamics}
\end{equation}
where $p_\com=[c_x,c_y,c_z]^T$ gives the CoM position, $M$ the total mass, and $a_g$ the gravitational acceleration \citep{Wieber2006a,Wensing2016}. It is important to note that $M \ddt p_\com = l_G$. As a result, these {\em centroidal dynamics} \eqref{eq:centroidal_dynamics} motivate solving a reduced problem posed over trajectories for the CoM $p_\com$, angular momentum $k_G$, contact locations $p_i$, and contact forces $\lambda_i$.

\begin{figure}
    \centering
    \includegraphics[width = .85 \columnwidth]{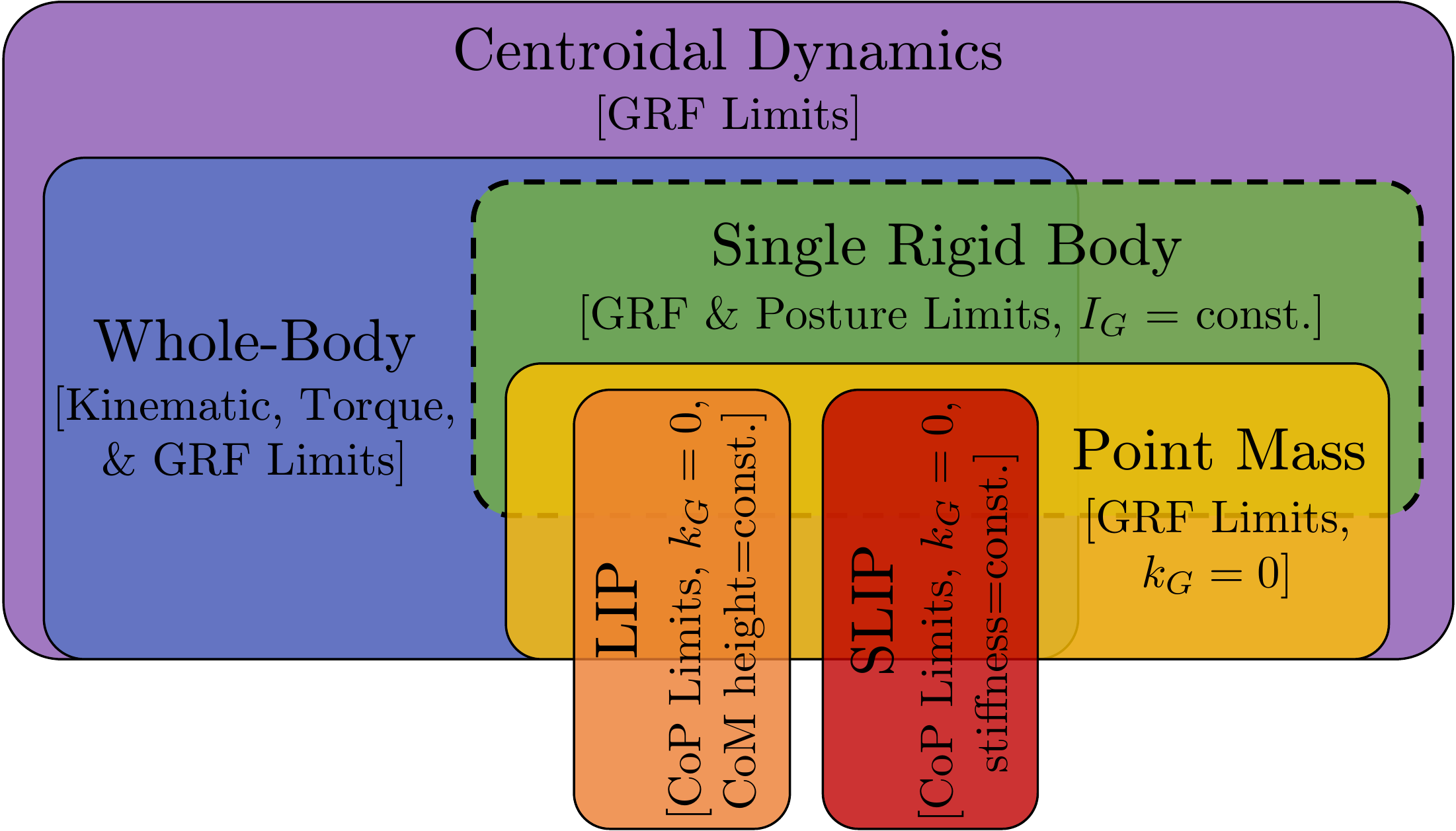}
    \caption{Relationship between motions feasible according to different models. Parentheses $[\cdot]$ list constraints imposed, which may be physical (e.g., GRF limits) or artificial (e.g., CoM height for LIP). For example, the strict containment of the whole-body model within the centroidal dynamics indicates that there exists feasible centroidal momentum trajectories for any whole-body feasible motion, but not vice versa (due to missing constraints in the centroidal model). Since the point-mass model imposes an artificial restriction $k_G=0$ but retains full GRF limits,  its set is strictly inside the centroidal one. LIP and SLIP relax full GRF feasibility by focusing on the CoP only, and so are not strictly within the point-mass region. Note: The size of the areas pictured does not convey the range of feasible motions in any precise sense.
    }
    \label{fig:motionFeas}
\end{figure}

While the reduction to a centroidal dynamics model addresses the high-dimensionality of the original problem, it remains non-trivial to ensure that centroidal solutions are whole-body feasible, since they neglect the geometric constraints and actuation bounds. This relationship is depicted in Fig.~\ref{fig:motionFeas} showing that only a subset of the motions that are feasible according to the centroidal dynamics are feasible for the whole-body model. 
This omission in the model has motivated planning with whole-body kinematics and centroidal dynamics \citep{Dai14}. Other strategies have separated these two, alternating between a centroidal dynamics and whole-body solve \citep{Herzog2016a,Budhiraja2019}.  Despite these advances, heuristic constraints on the CoM and footholds (e.g., to ensure reachability) remain prevalent, and simple bounds on the angular momentum (or fixing it to zero) are often used to simplify the problem. 

Even with these simplifications, the centroidal dynamics \eqref{eq:centroidal_dynamics} contains nonlinearities in the angular momentum equation due to bilinear terms from the cross product $(p_i-p_\com)\times \lambda_i$. Tailored methods to address these terms have been a topic of great recent focus. Dai et al.~\citet{Dai2016} considers fixing a-priori a polytopic or ellipsoidal region for the CoM, enabling optimization of an upper bound on the magnitude of the angular momentum. Valenzuela \citet{Valenzuela2016} considers relaxing the bilinear equalities with McCormick envelopes, which adds integer variables to the problem, but enables planning centroidal trajectories using mixed-integer convex optimization (see also, \citep{Ding2018}). 
Fernbach et al.~\citet{Fernbach2020} presented an exciting development where the contact locations are fixed and the CoM trajectory is parameterized via a B\'{e}zier curve with one free knot point, which avoids nonlinear effects.  
Overall, there are many choices for optimizing centroidal trajectories, with natural tradeoffs between flexibility, accuracy, and computational efficiency.

It is possible to further simplify the optimization of contact forces by considering their net effect. Given frictional limitations, each individual contact force $\lambda_i$ is constrained to a friction cone $\mathcal{C}_i$. Then, the net effect of these forces is a 6D cone known as the {\em Contact Wrench Cone} (CWC)
\begin{align}
CWC = &\left\{ \sum_{i=1}^{n_c} \begin{bmatrix} p_{i/\com} \times \lambda_i \\ \lambda_i \end{bmatrix} ~{\Big |}~ 
 \lambda_i \in \mathcal{C}_i,~\forall i \in \{1\, \ldots, n_c\} \right\} \nonumber.
\end{align}
When each friction cone is replaced with a polygonal approximation (e.g., friction pyramid), then the CWC is a polyhedral convex cone, and many tools from computational geometry are available to support its use in planning (c.f., \cite{Caron2015}). The CWC plays an integral role in checking for static stability in multicontact scenarios \citep{Bretl2008,Audren2018} and more recently has seen applicability for multicontact CoM motion generation \citep{Caron2016} and extension to address actuation limits \citep{Orsolino2018}. Feasibility constraints based on the CWC are natural generalizations of the celebrated zero-moment point (ZMP) criteria \citep{Hirukawa2006} to multicontact. Since the computation of the CWC is itself costly, it is most useful when the CWC can be precomputed offline (e.g., when contact locations are known in advance). 

\subsection{Other Simplified Models}

The centroidal dynamics \eqref{eq:centroidal_dynamics} are a special modeling simplification in that they represent a projection of the equations of motion without adding artificial restrictions on possible movements (i.e., any whole-body feasible motion can be projected to a feasible solution of the centroidal dynamics, as depicted in Fig.~\ref{fig:motionFeas}). 
Many other simple models follow by adopting artificial motion restrictions (or approximations) to further simplify analysis and optimization. 

A common simple model for quadruped locomotion is the single-rigid-body (SRB) model (e.g., \cite{DiCarlo2018a,ding2021representation,zhou2022momentum}). This simplification is motivated by the light leg designs of many quadruped robots, which makes the total rotational inertia of the system $I_G$ about the CoM approximately invariant with configuration \cite{Sim2022}. Thus, the SRB can be viewed as a restricted version of the centroidal momentum model in the case when the mass distribution takes a constant shape. Retaining a fixed inertia shape is not practically necessary, which makes this model an approximation in a way that simplified point-mass models are not. As a result, we denote this model slightly differently in Fig.~\ref{fig:motionFeas} in recognition of this distinction. Variations on this model have been proposed to accommodate inertia shaping \cite{lee2007reaction,zhou2022momentum}, however, there is an important subtlety that must be kept in mind in this case. While the linear momentum of the system is related to the CoM velocity (i.e., the rate of change in some average position), the angular momentum cannot, in general, be equated to the rate of change in any meaningful average orientation of the system overall \cite{saccon2017centroidal}. This property is intimately related to the fact that the conservation of angular momentum, in general, represents a non-holonomic constraint \cite{Wieber2006a}. 




The most common simple model for humanoid gait planning considers the centroidal dynamics with $k_G=0$ (i.e., thus removing consideration of orientation dynamics) and the additional restriction that the CoM height $c_z$ takes a fixed value denoted $h$. Rather than considering a collection of contact points, when walking on level ground, the overall Center of Pressure (CoP) position $p_\cop = [p_x,p_y,p_z]^T$ can be considered with $p_z=0$ an assumed ground height. These restrictions lead to the LIP Model \citep{Kajita2003} with dynamics:
$\ddot{c}_{x,y}  = \omega^2 ( c_{x,y} - p_{x,y} )$ 
where $\omega = \sqrt{g/h}$ represents the natural frequency of the LIP. The beauty and power of the LIP resides in the fact that these dynamics are linear, enabling convex optimization for planning CoM and CoP trajectories \citep{Herdt2010} or linear systems tools (e.g., LQR) for trajectory tracking \citep{Kuindersma2016}.

More recently, an area of interest has been discharging assumptions on the CoM height to enable more flexible motion plans \citep{Koolen2016}. One approach is to model height variations as a perturbation to the LIP \cite{Brasseur2015} along with constraint tightening to ensure robust feasibility. Rather than treating height variations as a disturbance, they can be introduced into the dynamics model through \citep{Feng2013, Koolen2016}. A common simple form is
$\ddot{p}_\com = a_g + k (p_\com - p_\cop)$  
where $k$ represents a variable stiffness-like parameter, constrained to be positive. 
Strategies for running (e.g., based upon the spring-loaded inverted pendulum model) \citep{Wensing2013f,Piovan2016} can be seen as placing additional restrictions on the form of $k$ based on assumptions of the virtual leg operating like a Hookean spring. While the SLIP dynamics are nonlinear, hybrid approaches can approximate SLIP physics via a combination of vertical spring dynamics and horizontal LIP dynamics \citep{Mordatch2010}, recovering linearity. Other hybrid approaches retain nonlinear dynamics \cite{Xiong2019,Xiong2020} by embracing more general NLP solvers. 

\subsection{Summary}
Modeling simplifications are commonly employed to reduce the computational burden for dynamic planning in legged robots. The most successful of these models (e.g., the centroidal model, SRB model, and LIP model) all focus their modeling detail on addressing the motion limitations imposed by contact interactions. The centroidal model doesn't impose any restrictions on motions (i.e., it represents a projection of the whole-body dynamics), while the other simplified models can be viewed as resulting from adding motion restrictions. These simplifications can result in additional structure (e.g., linearity) that can be leveraged when posing motion optimization problems over them (e.g., using the methods in the next section). In all cases, then there are many whole-body details that remain to be planned (e.g., swing foot trajectories between contacts). This omission represents a main downside to simple model planning, and has motivated the development of advanced numerical methods, as covered in Section \ref{sec:transcription}, for the efficient optimization of whole-body plans. 


\begin{table*}[t]
    \centering 
    \def\arraystretch{1.4}
    \begin{tabular}{ c  c   c  c } 
        \hline 
         & & {\bf \em Supporting Theory} & {\bf \em Computation}  \\\hline
        {\bf Global} & &  Hamilton-Jacobi-Bellman (HJB) & Numerical PDE Solvers or Dynamic Programming \\\hline 
        \multirow{2}{*}{\bf Local} & {\bf \em Direct} {(\em discretize then optimize)} &  Numerical Integration & \bf{Direct Shooting/Collocation} \& NLP \\ 
        & {\bf \em Indirect} ({\em optimize then discretize}) & Pontryagin Maximum Principle (PMP) & Indirect Shooting/Collocation \& Root Find 
     \\\hline
    \end{tabular}
    \vspace{5px}
    \caption{High-level relationship between optimal control solution methods. This survey focuses mostly on direct methods for continuous-time problems due to their prominence in robotics applications.}
    \label{tab:OC_Methods}
\end{table*}

\subsubsection{Relationship with learning}
There are many ways in which emerging learning strategies have been used to advance the application of simplified models for locomotion planning. With regards to the models themselves, this review has covered existing strategies based on considerations of physics and expert intuition. Moving forward, the automatic discovery of simple models remains an important open problem (c.f., \cite{Chen2020}), potentially leveraging a range of dimensionality reduction tools from the learning literature. Other work has looked at how to close the gap between these simplified models and their whole-body counterparts, for example, to learn constraints on simplified models that address kinematics constraints at the whole-body level \cite{Carpentier2017} or that address the gap via learning robust control strategies for MPC \cite{pandala2022robust}. Yet other strategies have wrapped full RL pipelines around low-level control based on simple models \cite{xie2021glide} in an effort to increase the sample efficiency of learning. It should be acknowledged that even though RL strategies are often architected (e.g., via the autoencoder trick) to learn low-dimensional representations within their layers (c.f.,~\cite{lee2020learning}), these latent representations remain difficult to interpret. It is an open problem whether the existing simplified models may be used to increase the interpretability of learned controllers as well.  




\section{Numerical Methods for Solving OCPs}
\label{sec:transcription}


While the decomposition of our general problem formulation in Section~\ref{sec:formulation} (e.g., via modeling simplifications) remains a bit of an art, the technical details of solving a trajectory optimization problem represents a field unto its own. This field of dynamic optimization  has benefited from contributions across engineering applications ranging from chemical process control \cite{biegler2009large}, to flight planning \cite{malyuta2022convex}, and others. We begin this section by making a few simplifications to our general problem \eqref{eq:ocp_wb}, so that we may review  the most common methods for solving OCPs at large, which provides context for the most common approaches taken for robotics in particular. We then gradually discharge our original simplifications, and explain nuaunces that arise in OCPs for locomotion. The treatment of these nuances has led to a rapid increase in whole-body optimization in recent years, 
where the need for fast online re-planning has led the community to 
develop fast structure-exploiting solvers tailored for robotics.




To begin, let us consider \eqref{eq:ocp_wb} in the case of a single contact phase (e.g., optimizing a trajectory for a humanoid doing an in-place dance on two-feet). For better alignment with the dynamic optimization literature, we rewrite the dynamics \eqref{eq:ocp_wb_dyn} as a system of first-order ODEs $\dot x(t) =  f( x(t), u(t), p(t) )$ with time-varying parameters $p(t) = \lambda(t)$. The formulation can also be readily extended to consider time-invariant parameters. In the presence of contacts, the contact points must not move relative to the ground, which can be written as an algebraic constraint $g(t, x(t), u(t), p(t)) =0$. These ODEs, combined with the algebraic constraints, lead to system dynamics that are more formally classified as a system of differential-algebraic equations (DAEs) \cite{biegler2009large}. A generic nonlinear optimal control problem for such a system can then be considered as:
\renewcommand{\l}{\ell}
\newcommand{\ph}{{\rm ph}}
\begin{subequations}
\begin{align}
\minimize_{x(\cdot),u(\cdot),p(\cdot)} ~~& \int^{t_f}_{t_0} \l(x(t), u(t), p(t)) dt + L(x(t_f))
\label{eq:ocp_gen_f} \\
{\rm subject~to}~~
& \dot x(t) =  f( x(t), u(t), p(t)) \label{eq:ocp_sub:f}\\
& 0 = g(x(t), u(t), p(t))  \quad \label{eq:ocp_sub:g}\\
&~~~~~~~~~~~~~~~~~~~~\forall t \in [t_0, t_f] \nonumber\,,
\end{align}
\label{eq:ocp_gen_c}
\end{subequations}
where the objective function, with running cost $\l$ and  terminal cost $L$, is minimized over the time interval $[t_0, t_f]$ subject to the system dynamics  defined by the DAEs. 

For locomotion problems, the formulation \eqref{eq:ocp_gen_c} requires extension to consider how the dynamics change in different contact modes. Here, we assume that the mode sequence is fixed a-priori, and consider the multi-phase problem: 
\begin{subequations}
\begin{align}
\minimize_{\substack{x(\cdot), u(\cdot),p(\cdot), \\[.5ex] s_1,\ldots,s_{n_{\ph}}}} ~~&  \sum_{j=1}^{n_{\ph}} \left[ \int^{s_j}_{s_{j-1}} \l_j(x(t), u(t), \lambda(t)) dt + L_j( x(s_j)) \vphantom{\int^{s_{j+1}}_{s_j}} \right]
\label{eq:ocp_gen_mp}\\
{\rm subject~to} ~~ &
\dot x(t)  =  f_j( x(t), u(t), \lambda(t)) \label{eq:ocp_mp_sub:f}\\
&g_j(x(t), u(t), \lambda(t)) = 0 \quad \label{eq:ocp_mp_sub:g}\\
& h_j( x(s_j^-)) = 0 \label{eq:ocp_mp_sub:guard} \\
&x(s_j^+) = R_j(x(s_j^-)) \label{eq:ocp_mp_sub:x0} \\
& ~~ \forall t \in [s_{j-1}, s_{j}], ~ j = 1,...,n_{\ph} \\
& s_0=t_0, s_{n_{\ph}}=t_f\,,
\end{align}
\label{eq:ocp_mp_sub}
\end{subequations}
where $j=1,...,n_{\ph}$ denotes the phase index, with the time interval of the $j$-the phase given by $[s_{j-1}, s_{j}]$, and $n_{\ph}$ the number of phases. 
At the mode switch, a guard constraint \eqref{eq:ocp_mp_sub:guard} is typically used to ensure the active contact points are in contact with the environment. Depending on the contact model, the reset maps \eqref{eq:ocp_mp_sub:x0} may correspond to the impact dynamics or to switching between coordinate representations. 

Generally speaking, many methods for solving an OCP only apply directly to \eqref{eq:ocp_gen_c} in the ODE case, but often admit extensions to the DAE and multi-phase settings. These methods can be categorized in three groups: global methods based on the Hamilton-Jacobi-Bellman (HJB) equation (or Dynamic Programming in discrete-time), local indirect methods, and local direct methods (see Table~\ref{tab:OC_Methods}). 

Methods based on Dynamic Programming exploit Bellman's principle of optimality~\cite{bellman1954theory} to solve a discrete-time version of the OCP~\eqref{eq:ocp_gen_c}. They do so via finding the optimal cost-to-go $V(t_0, x_0)$ (also called the \emph{Value function}). 
In continuous time, the Value function must satisfy a partial differential equation (PDE) known as the HJB. 
Its solution can be approximated by discretizing time and space and applying dynamic programming (DP). However, it is well-known that the complexity of this strategy increases exponentially with the number of states and controls. Therefore, it is not directly applicable to most legged robots, requiring clever approximate decomposition strategies for application to whole-body models~\cite{whitman2010control}. 

Indirect methods transform the original OCP into a Boundary Value Problem (BVP) by using Pontryagin’s Maximum Principle (PMP) \cite{liberzon2011calculus} to formulate the so-called co-state equations. This approach enables pre-optimizing the control as a function of state and co-state if the dynamics is control affine---which is the case for most legged robots. Indirect methods turn the OCP into a root finding problem, which  can provide fast and accurate solutions, but suffers from high sensitivity to the initialization. 
These methods are challenging to initialize, 
and it is also not possible to directly enforce state inequality constraints. 

Direct methods, on the other hand, have been more broadly adopted by the legged robotics community for motion generation, and are generally compatible with systems of DAEs. 
In direct methods, the OCP is transcribed into a finite-dimensional Nonlinear Program (NLP) by discretizing the controls and states with respect to time. Therefore, direct methods directly find the minimum of \eqref{eq:ocp_cost}, where the NLP can be solved with well-established optimization techniques, e.g., Sequential Quadratic Programming (SQP)~\cite{wright1999numerical}.
The difference between direct methods lies in the way the discretization is carried out. 
In the following, we focus exclusively on the three prominent direct methods in the community in recent years, namely \emph{direct multiple shooting}, \emph{direct collocation}, and \emph{Differential Dynamic Programming (DDP)}. A comprehensive review of numerical methods for trajectory optimization (including discussion of indirect shooting and indirect collocation) can be found in \cite{betts2010practical}. 

\subsection{Multiple shooting} 
\begin{figure}[!tbp]
   \centering
   \includegraphics[width=.85 \columnwidth]{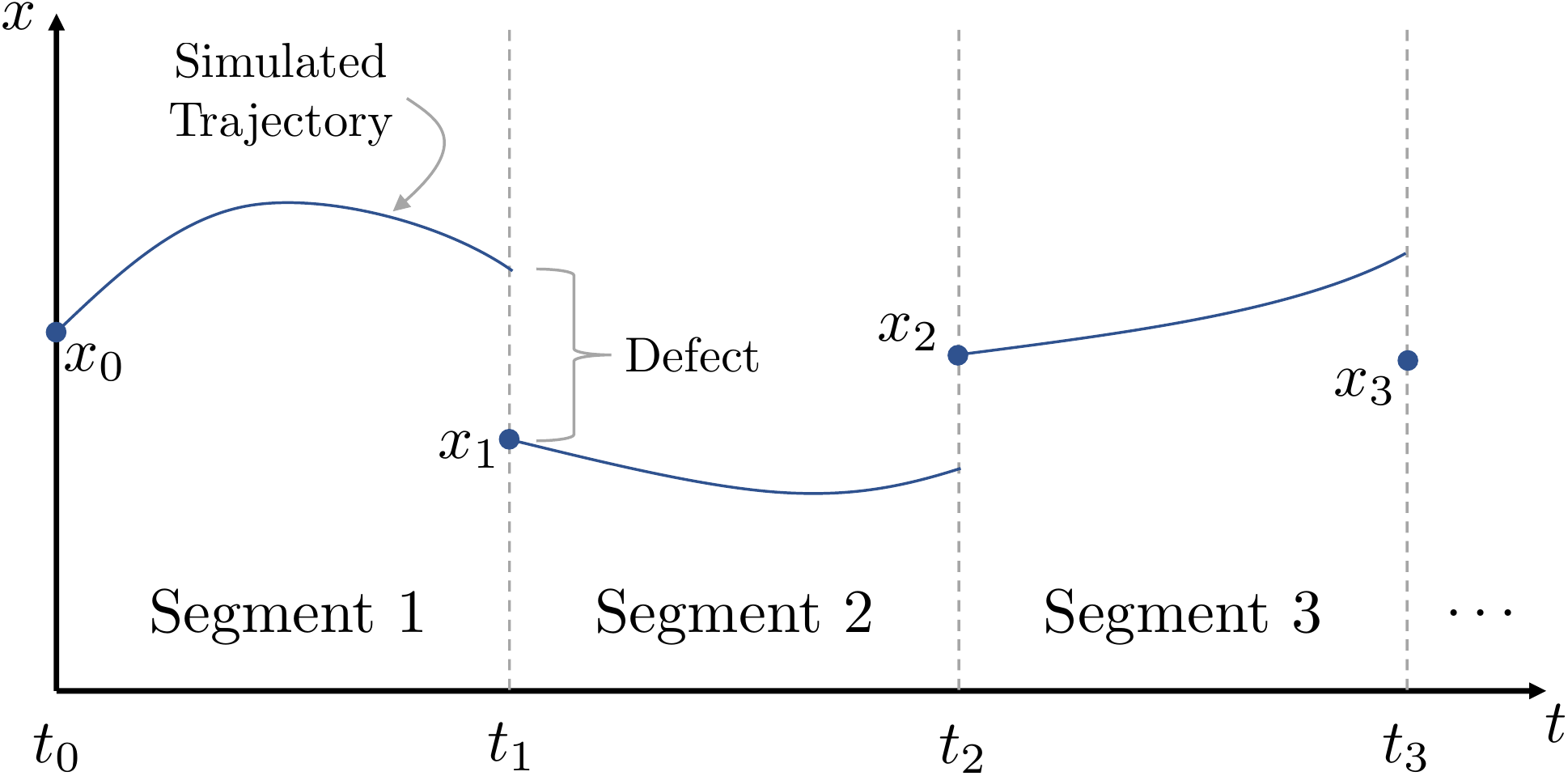} 
   \caption{Depiction of multiple shooting. Each shooting segment is associated with a numerical simulation (e.g., via a numerical integration method). Optimization variables include the states $x_i$ at the beginning of each segment and variables describing the controls applied in each segment. Constraints enforce continuity, i.e., that the defect between segments is zero.}
   \label{fig:multipleshooting}
\end{figure}
In direct shooting methods, the controls $u(t)$ are discretized over the time horizon, while the state trajectories are obtained via forward integration. 
In contrast to \emph{single shooting}, 
where a single integration is performed over the whole time horizon, in multiple shooting, the time horizon is discretized using a finite grid, then integration is performed on each of the segments (Fig.~\ref{fig:multipleshooting}), i.e., at each multiple shooting node. The initial state of each segment is added to the decision variables, and \emph{continuity constraints} are introduced to guarantee that the final state of each segment $i$ (computed with integration) matches the initial state of the next segment $i+1$. In essence, as in Fig. \ref{fig:multipleshooting}, the initial optimization problem is divided into smaller optimization problems over the discretized grid, whose initial conditions can be set separately. By parameterizing the time horizon with a series of Initial Value Problems (IVP), the time interval over which integration is performed is shortened, reducing the high sensitivity issue of single shooting and preventing states from diverging. This makes multiple shooting a method that can deliver rather robust solutions as shown in \cite{bock1984multiple}, which motivated many researchers to use it in the context of motion generation for complex robots. 

In particular, the use of multiple shooting has been popular for generating motions considering whole-body dynamics where contact modes are set a priori, i.e., predefined contact sequences. This is the case of~\cite{Mombaur2009}, \cite{schultz2009modeling}, and \cite{koch2014optimization}, where contacts are assumed to be rigid and impacts are instantaneous and inelastic. The resulting OCP is then a multi-phase problem with discontinuous phase transitions, as in~\eqref{eq:ocp_mp_sub}. 
In \cite{hereid2015hybrid}, the authors used multiple shooting to optimize virtual constraints, coupled with Hybrid Zero Dynamics (HZD), for the gait generation of a compliant bipedal robot.


\subsection{Collocation} 
\begin{figure}[!tbp]
   \centering
   \includegraphics[width=.85\columnwidth]{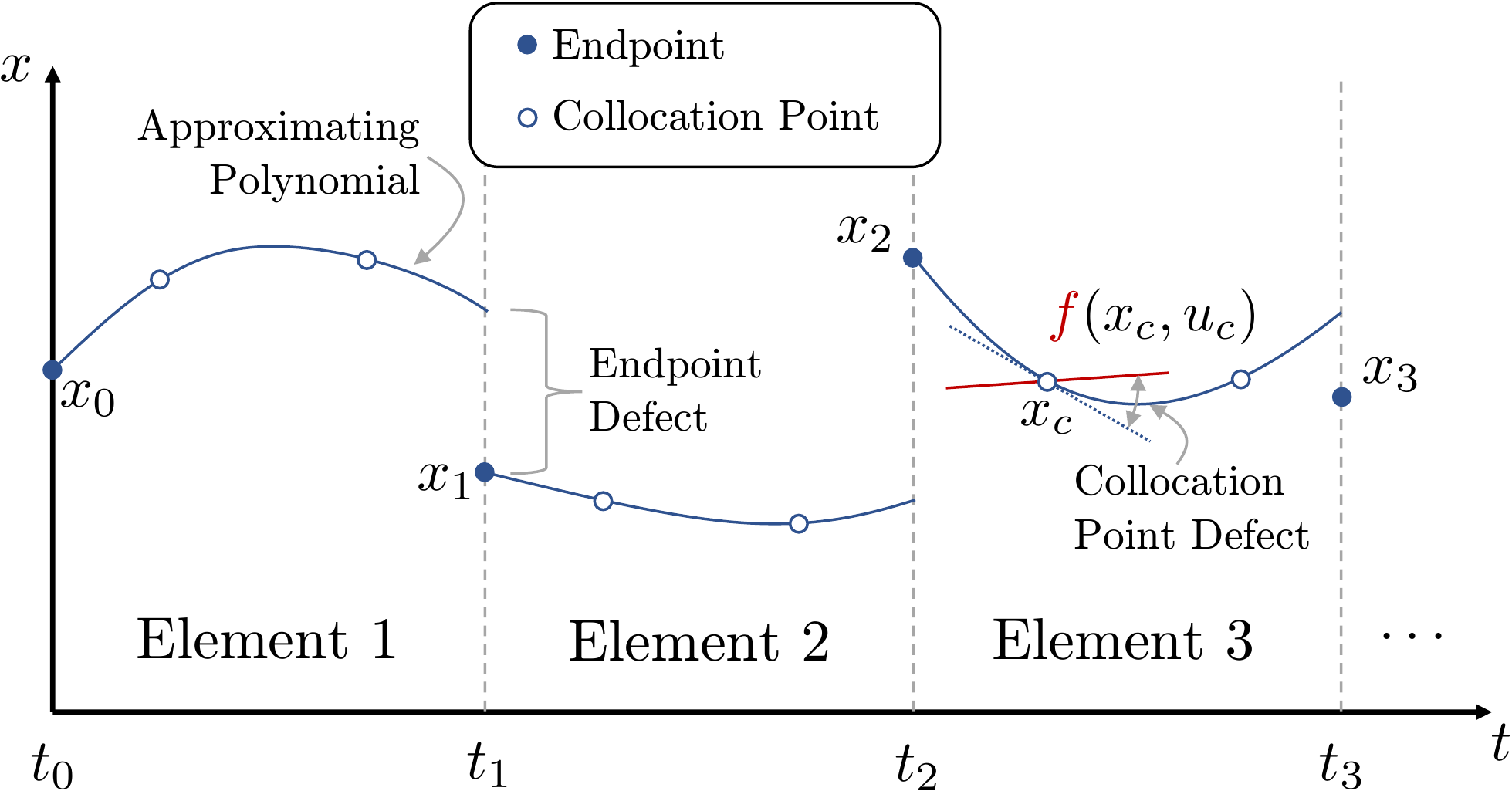} 
   \caption{Depiction of collocation. Each finite element is associated with a polynomial that approximates the trajectory. Optimization variables include the states $x_i$ at each endpoint, and states/controls at each collocation point. Constraints enforce that the endpoint defect is zero, and that the slope of the approximating polynomial matches the evaluation of the dynamics function $f$ at each collocation point. }
   \label{fig:collocation}
\end{figure}
In direct collocation (also known as direct transcription \cite{betts2010practical}) the original OCP is transformed into an NLP having control and state trajectories as decision variables. Controls and states are discretized over a time grid, where the intervals are called \emph{finite elements} (see Fig.~\ref{fig:collocation}). 
Controls are approximated in each finite element by a finite dimensional representation, and the states are approximated by polynomials. The system dynamics is then enforced imposing {\em continuity constraints} to eliminate endpoint defects at the end of each element and though internal \emph{collocation constraints} imposed at each of the \emph{collocation points} (see Fig.~\ref{fig:collocation}). These collocation constraints enforce that the slope of the approximating polynomial matches the system dynamics $f( x(t), u(t), p(t))$ at each collocation point. 
The choice of the number and location of these collocation points has been thoroughly investigated~\cite{Hager2016} because it affects the accuracy of the method, namely how closely the NLP approximates the original OCP.
More details on collocation methods can be found in \citep{hargraves1987direct, von1992direct,Kelly2017}.

Collocation has been a popular method to compute walking motions for legged robots using whole-body models and contact constraints. In \cite{Hereid16}, a periodic motion was generated for a whole-body bipedal robot, where the convergence of the periodic cycle is guaranteed through the implementation of HZD virtual constraints. 
In \cite{pardo2017hybrid}, the authors used collocation to generate motions for a quadruped robot, where the contact sequences were predefined.
Collocation constraints were modified for solutions to respect contact constraints in \cite{Posa2016}, providing third-order integration accuracy with demonstrations on a full-body humanoid. 

Most robotics methods to date have considered the above formulation where the system dynamics are transformed into a system of first-order ODEs. Vanilla collocation methods apply separate approximating polynomials for each component of the state (e.g., separate polynomials for $q$ and $\dot{q}$), with their consistency only enforced at collocation points \cite{Moreno2022collocation_second_order}. Recent work \cite{Moreno2022collocation_second_order,hoang2021collocation} has shown the benefit of using a unified interpolating polynomial for the components of $q$ and $\dot{q}$, which ensures their consistency across the trajectory, 
and improves the dynamic accuracy of the solution \cite{Moreno2022collocation_second_order}. 


\subsection{Differential Dynamic Programming (DDP)} 
\label{sec:ddp}
In recent years a high interest has grown towards DDP and its variants. DDP~\cite{mayne1966second} is an age-old method for solving discrete-time unconstrained OCPs. The method is based on Dynamic Programming, but it overcomes the curse of dimensionality by working on a local quadratic expansion of the so-called ``Q function'' (the sum of the current stage cost and the Value function at the next state). This makes DDP different from Dynamic Programming, because the method does not attempt to exhaustively explore the state space of nonconvex problems. DDP roughly can be seen as an efficient iterative algorithm for solving the banded system of linear equations associated with the KKT conditions of an unconstrained OCP transcribed with collocation. Indeed, DDP closely resembles Netwon's method~\cite{Murray1984}; it has the same convergence rate (quadratic), but it is not completely equivalent to it. The main difference is that when solving the KKT system, DDP expresses the control inputs as linear functions of the state, but then, instead of computing control perturbations by using the linearized dynamics from the KKT system, it performs a forward simulation with the original nonlinear dynamics. This makes DDP a single shooting strategy---because the state is computed by integrating the dynamics---but it is better suited to handle unstable dynamics compared to vanilla shooting methods since the feedback form of the optimal control inputs helps to prevent divergence.

DDP requires the second derivatives of the dynamics to achieve quadratic convergence. While this is a nice property, it may be challenging to compute these terms for complex systems. So, in recent literature, variants of DDP that make use of Hessian approximations have been more popular, such as iLQR and iLQG \cite{Todorov2005,Tassa2012, Geisert2017}, with other recent work focused on efficiently computing the full Hessian \cite{nganga2021accelerating,singh2022analytical}. 


The main limitation of DDP is that, in its original form, it cannot handle constraints besides the system dynamics. However, DDP has been extended in many ways to overcome this issue. Box DDP~\cite{Tassa2014} is a simple extension that accounts for box constraints on the control inputs. Hierarchical DDP~\cite{Geisert2017} allows for the optimization of a hierarchy of cost functions in lexicographic order. DDP extensions that can treat arbitrary nonlinear inequality state-control constraints have been proposed based either on interior-point techniques~\cite{Pavlov2020}, active-set methods~\cite{Xie2017}, Augmented Lagrangian~\cite{Howell2019}, or relaxed barrier \cite{Grandia2019a} approaches. Recently DDP has also been extended to handle implicit dynamics~\cite{Chatzinikolaidis2021}, hybrid dynamics~\cite{Li2020hybrid}, and multi-phase dynamics~\cite{Budhiraja2019a}, all of which are useful for handling contacts. Even though DDP was born as a single-shooting method, it has been extended to multiple shooting~\cite{Pellegrini2017,plancher2018performance,giftthaler2018family} to further improve its handling of sensitive dynamics.


\subsection{Contact Implicit Considerations}

As a departure from the main assumptions of this section, contact-implicit (or contact-invariant) formulations (e.g., as first discussed in Section \ref{sec:contact:hybrid_seq_optimization}) instead seek to optimize the mode sequence. The choice of transcription strategy is highly coupled with the choice of contact model adopted. When using shooting methods, the key need is to be able to provide gradients that relate changes in states/controls to trajectory outcomes. For example, when adopting a relaxed contact model, the outcomes of simulation are always differentiable, enabling the use of shooting solvers and/or DDP \cite{Tassa2012,Chatzinikolaidis2021}. 
When adopting an event-driven hybrid model, these methods can still be used when paired with suitable hybrid sensitivity analysis, as in \cite{Kong2021}. Likewise, discrete-time shooting can be applied with time-stepping LCP solvers \cite{Carius2018}, although both this case and the hybrid case exhibit pathologies that prevent differentiability in corner cases (c.f.,~ Sec.~\ref{sec:hybrid_differentiability}).


When using direct collocation methods, existing contact-implicit methods rely on the inclusion of complementary constraints in the optimization. First-order accurate methods in \cite{Posa2014} have been extended to allow for high-order schemes assuming the mode is not changing during an element  \cite{xi2016selecting,Manchester2019a,Patel2019}. These methods either require impacts at the element boundaries \cite{Patel2019,shield2022contact} or allow for impacts during the element at the expense of integration accuracy \cite{Patel2019,Manchester2019a}. There is a clear trade-off here, since enforcing impacts at the element boundaries requires additional complementarity constraints, which are notoriously difficult to enforce. The alternative is to consider relaxations of the LCP constraints \cite{Cleach2021}, which again present trade-offs for accuracy and computational complexity.



\subsection{Summary}

Within robotics, direct methods remain the most attractive numerical approaches for trajectory optimization. While both direct shooting and direct collocation can be used in cases when the mode sequence is fixed a priori, contact-implicit strategies require more careful consideration regarding the contact modeling choices adopted and how those affect the available transcription approach. Recent years have seen progress on accelerating the solution of shooting problems through the use of DDP, with recent sparse QP solvers (e.g., \cite{frison2020hpipm}) opening the door for similar accelerations to closely related collocation formulations. 

\subsubsection{Relationship to learning}
There have been many motivating examples of learning being used to support trajectory optimization. 
A common strategy is to employ learned networks to provide either a policy or a trajectory to warm start optimization (see, e.g., \cite{melon2021receding, mansard2018using}). Another common way for learning to support trajectory optimization is by learning the value function, which enables online trajectory optimization over shorter horizons \cite{lowrey2018plan}. Other strategies simplify the optimization by learning footholds \cite{gangapurwala2022rloc}, e.g., akin to a higher-level contact planner.

\section{Realizing Motion Plans}
\label{sec:control}



\begin{figure}
    \centering
    \includegraphics[width=\columnwidth]{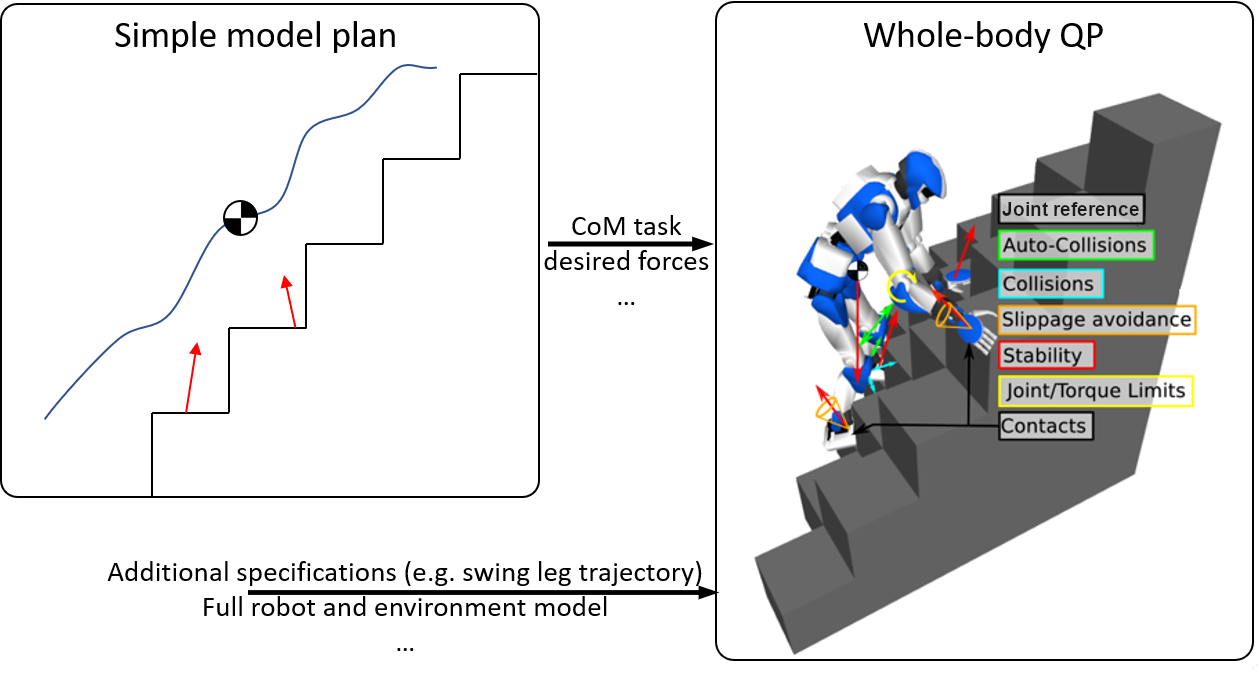}
    \caption{A classical use of a Whole-Body QP in a pipeline to execute a simple-model plan obtained from the methods of the previous sections. A typical list of tasks and constraints is also depicted.}
    \label{fig:QPControl}
\end{figure}

When using the tools of the previous section (posed over either simplified or whole-body models) for real-time predictive control, there often arises a need for separate reactive control components to realize the motion plans created. For example consider the setup depicted in Fig.~\ref{fig:QPControl}. When performing trajectory optimization over simplified models (e.g., the SRB or a point-mass model), the OCP may be able to be solved quickly, but there still remain many details that need to be defined (e.g., swing leg motion). For trajectory optimization over whole-body models, the solver is often limited to running at a slower rate (though these rates are continually improving), requiring some additional high-rate closed-loop control. Optimal closed-loop feedback policies can sometimes be extracted from TO solutions, e.g., as with DDP \cite{Grandia2019a}. In most other cases, some additional reactive control is required to realize optimized motion plans and handle disturbances.

Conceptually, when the time interval in the OCP \eqref{eq:ocp_wb} is reduced to a single instant, the OCP becomes a problem of \emph{instantaneous control}, with the functional decision variables and their derivatives becoming simple vector variables, and objectives and constraints considering a single point in time.
While it is conceptually interesting to consider instantaneous control schemes as limiting cases of the OCP, this is not how they were developed. Rather, the topic stands on its own, with a large body of literature and schemes that are widely used in practice on legged robots, either as a direct way to generate motion, or as a means to execute trajectories generated online with simplified models or offline by whole-body OCP.

The popularity of instantaneous reactive control schemes owes to their relative ease of implementation and the fact that, with a few assumptions and well-chosen objective/constraint formulations, the problem can be written as a Quadratic Program (QP), or a close variant, and solved very quickly (from hundreds of microseconds to a few milliseconds).

\subsection{Formulation}
In instantaneous settings, one common way to define the motion is through quantities $e_i$ that we want to regulate to $0$, or to keep above $0$. For example, let us suppose that a walking gait has been generated using the LIP model, and we want the CoM of the robot to follow the LIP plan. For the robot CoM to track that trajectory, we want to achieve $e_i(q,t)=0$ where $e_i \in \mathbb{R}^3$ gives the error between the target position from the LIP and the true position deduced from $q$ by forward kinematics. Other tasks, such as collision avoidance, are naturally captured by inequality constraints $e_i(q) \geq 0$ with, for example, $e_i(q)$ being the distance between a robot body and an object.
More generally, $e_i$ can be a function of any subset of $(q,\qdot,t)$. Such a definition is often called a \emph{task error function} by reference to the task function~\cite{Samson1991} or operational-space~\cite{khatib1987unified} formalism, but can be also considered as an output of the system from a more controls-centric point of view.

A key point is that the dynamics \eqref{eq:ocp_wb_dyn} are linear in $\qddot$, $\tau$, and $\lambda$. This points at writing the regulation at the acceleration level where $\tau$ and $\lambda$ can also be used directly to affect motions. 

\subsubsection{Task Dynamics}
Consider a general task, in the special case where $\dot{q} = \qdot$ and $e_i \in \mathbb{R}^m$. Differentiating $e_i(q,t)$ twice with respect to time gives
\begin{equation}
	\ddot{e}_i(q,\qdot, \qddot, t) = J_i(q) \qddot + \dot{J}_i(q,\qdot) \qdot + a_i(q,\qdot,t)\,,
\end{equation}
where $J_i$ is the Jacobian matrix of $e_i$ and $a_i = \dfrac{\partial^2 e_i}{\partial t^2}$. This construction can also be generalized to the case where $q$ and/or $e_i$ do not lie in Euclidean space (e.g., for orientation tasks). 

To bring or keep $e_i$ to a desired value, one then writes $\ddot{e}_i = \ddot{e}_i^d$ or $\ddot{e}_i \geq \ddot{e}_i^{min}$ where $\ddot{e}_i^d$/$\ddot{e}_i^{min}$ is a function of any subset of $(q,\qdot,t)$ that indicates how we want to regulate $e_i$ (see \ref{sec:control:regulation}).
This gives an equality/inequality that is linear in $\qddot$:
\begin{equation}
	J_i(q) \qddot \geqq \ddot{e}_i^d - \dot{J}_i(q,\qdot) \qdot - a_i(q,t)\,. \label{eq:linearized_task_q}
\end{equation}
If $e_i$ depends on $\qdot$, a single differentiation leads to:
\begin{equation}
	\dfrac{\partial e_i}{\partial \qdot} \qddot \geqq \dot{e}_i^d - J_i(q,\qdot) \qdot - \dfrac{\partial e_i}{\partial t}(q,\qdot, t)\,. \label{eq:linearized_task_nu}
\end{equation}
Both \eqref{eq:linearized_task_q} and \eqref{eq:linearized_task_nu} can be written as
$
	A_i(q, \qdot) \qddot \geqq b_i(q, \qdot, t) \label{eq:task_constraint}
$\,.

\subsubsection{Task expressions}
The most common tasks involve geometric features (e.g., body or CoM position) that are direct functions $f_i$ of $q$. 
Typical equality tasks have the form $e_i(q,t) = f_i(q) - f_i^d(q,t)$, where $f_i^d$ is the target value. For non-Euclidean tasks, this error can be generalized as $e_i(q,t) = f_i(q) \ominus f_i^d(q,t)$, where $\ominus$ denotes a meaningful difference in the task space \cite{sola2018micro}.
This general definition can capture tasks for the CoM position, body positions, body orientations, posture, and others. An important use of position and orientation tasks is to describe the geometric part of a contact. Other equality tasks include a gaze task\cite{Saab2013}, visual servoing\cite{Agravante2017}, 
or any subpart of one of the above\cite{Feng2015,Vaillant2016}. 
Geometric inequality tasks commonly include bounds, collision avoidance, balance, etc.~\cite{Audren2016,Saab2013}.

Few tasks depend directly on $\nu$. This is obviously the case of joint speed limits $\underline{\qdot} - \qdot \geq 0$ and $\overline{\qdot} - \qdot \geq 0$, but also the centroidal momentum task\cite{DeLasa2010,Wensing2013g,Herzog2016} $e_i = A_G(q)\qdot$ from \eqref{eq:CMM}.

We can also call tasks, by extension, constraints or motion requirements that are directly written on $\qddot$, $\tau$ and $\lambda$ and do not need regulation. These include, for example, limits on $\qddot$, $\tau$, as well as the friction constraints on the forces (to prevent sliding), that are usually approximated using a linear form $C \lambda \geq 0$ (e.g., as was also considered to simplify the CWC in Sec.~\ref{sec:simplified_models:centroidal}). Direct references can also be given for these variables. For example $\lambda-\lambda^d=0$ can be used for force control\cite{Bouyarmane2019}.

\subsubsection{Regulation of Equality Tasks}
\label{sec:control:regulation}
The behavior the controller must achieve to keep a task error $e_i$ at or above $0$ can be described by a desired value for its derivative $e_i^{(l)}$ where $l$ is the number of differentiations (0, 1 or 2) such that $e_i^{(l)}$ is an affine function of $\qddot$, $\tau$ or $\lambda$.
For geometric equality tasks ($l=2$), the most common regulator is a PD construction: $\ddot{e}_i^d = -K_d \dot{e}_i  -K_p e_i$, where $K_d$ and $K_p$ are positive definite matrices, which are usually taken as diagonal or scalar matrices to get fully decoupled dynamics. $K_d$ is typically set to $2K_p^{1/2}$ to obtain a critical damping.
Constant gain matrices work well when tracking a trajectory or when $e_i$ remains small, but incur either large acceleration or slow convergence when $e_i$ starts large. Solutions to work around this issue include gains varying with the norm of $e_i$ (e.g., main ideas in \cite{Abe2007, Saab2013}), or clamping/scaling $\ddot{e}_i^d$. 

A common (and critical) equality task is the one related to maintaining contacts. Most often, hard contacts are assumed and $\ddot{e}_i^d$ is set to $0$, even though it may be better to set a damping term to stabilize the contact in case of slippage~\cite{Vaillant2016}. As a related aside, it should be noted that around the initiation of contact events, this constraint can lead to a rapid change in the problem formulation and thus a rapid change in controls. 

For velocity-level equality tasks, regulation is typically achieved with a simple linear feedback: $\dot{e}_i^d = K_p e_i$.
For acceleration/force-level equality tasks, regulation is either omitted (e.g., $e_i^d=0$), or implemented with Proportional-Integral schemes: $e_i^d = -K_p e_i -K_i \int e_i \, \text{d}t$.


An alternative for making $e_i$ converge to zero is to view the task error $e_i$ as an output of the system and to design a control Lyapunov function $V_i$ stabilizing it to $0$. Then the inequality $\dot{V_i} \leq -\gamma_i V$, where $\gamma_i$ sets the desired exponential convergence rate, forces the convergence of $e_i$ to $0$. This is the idea behind the CLF-QP \cite{Ames2013}, which brings convergence guarantees provided the inequalities are always feasible. However, ensuring persistent feasibility with practical robot models remains an open challenge.

\subsubsection{Regulation of Inequality Tasks}
\label{sec:control:regulation:ineq}
Inequality tasks at the acceleration level, such as joint torque limits or force friction cones, do not require any regulation and can be directly imposed.
Inequality tasks at the velocity level, such as joint velocity limits, can be easily regulated by computing $\dot{e}_i^d$ such that $e_i$ remains positive at the next time step: $e_i^+ = e_i + \Delta t \, \dot{e}_i \ge 0$, where $\Delta t$ is the time step of the controller. This approach works well for small $\Delta t$ (which is usually the case), and implies setting $\dot{e}_i^d = -e_i / \Delta t$.
Inequality tasks at the position level, such as joint position limits or obstacle avoidance, are much harder to enforce. The most common approach is to compute $\ddot{e}_i^d$ so that $e_i^+$ is positive, but this can easily lead to conflicts with the acceleration-level constraints~\cite{Park1998}. This has motivated the development of several alternative approaches, ranging from the simple trick of using a larger value of $\Delta t$ in the computation of $e_i^+$~\cite{Park1998,Saab2013}, to bounding velocities with a hand-tuned function of the distance to the position limit~\cite{Kanehiro2010}, to control barrier functions~\cite{Nguyen2016, Rauscher2016}. However, none of these approaches ensure compatibility between position-level and acceleration-level inequalities. The only methods that can guarantee them are those that account for position-level and acceleration-level inequalities at once~\cite{Decr2009,Rubrecht2012,DelPreteRAL2018}, even if they only work under the assumption of constant acceleration bounds, which is not the case in practice. 
%
%
Fundamentally, enforcing position constraints remains precarious with instantaneous control, since it is the evolution of the system dynamics over an extended time period that determines whether the position constraint can be respected in the future. This observation motivates the need for predictive strategies (e.g., repetitively solving trajectory optimization formulations) in order to address position inequality constraints via suitable lookahead.

\subsection{Resolution}
Instantaneous control schemes make the assumption that the state $x$ is known and constant during the control period $\Delta t$. This is supported by the fact that $\Delta t$ is small (typically between 1 and 10 ms), and relies on the ability to solve the problem in that time. The contact state (i.e., which bodies are in contact) is assumed given and fixed at each control cycle. 

Under these conditions, and if the friction cones are approximated as pyramids, aggregating all the desired behaviors of the tasks leads to a set of linear constraints:
\begin{subequations}
\label{eq:instantaneous_problem}
\begin{align}
	  \makebox[0pt][l]{$M$}\phantom{A_{\qddot,i}} \qddot - \makebox[0pt][l]{$J^T$}\phantom{A_{\lambda,i}} \lambda - \makebox[0pt][l]{$S^T$}\phantom{A_{\tau,i}} \tau &= -C\qdot - \tau_g \label{eq:ocp_wb_dyn2} 
	  \\
	  \makebox[0pt][l]{$\makebox[0pt][l]{$J_c$} \phantom{A_{\qddot,i}} \qddot$} \phantom{A_{\qddot,i} \qddot + A_{\lambda,i} \lambda + A_{\tau,i} \tau} &= a_c \label{eq:contact_tasks}
	  \\
	  A_{\qddot,i} \qddot + A_{\lambda,i} \lambda + A_{\tau,i} \tau  &\geqq b_i \quad \mbox{for non-contact tasks $i$}\,,\label{eq:non_contact_tasks}
\end{align}
\end{subequations}
where \eqref{eq:ocp_wb_dyn2} is a  reorganization of \eqref{eq:ocp_wb_dyn}, \eqref{eq:non_contact_tasks} is a generalization of \eqref{eq:task_constraint} to account for tasks directly written on $\tau$ or $\lambda$, and we have singled out the contact constraints in \eqref{eq:contact_tasks}. 

Problem \eqref{eq:instantaneous_problem} represents what we would like the robot to achieve. The idea is to solve \eqref{eq:instantaneous_problem} for $\dot{\nu}$, $\tau$, and $\lambda$ as best as possible (for some definition of best given below). On position- or velocity-controlled robots, $\qddot$ is integrated twice or once and the resulting configuration~\cite{Vaillant2016} or velocity~\cite{Agravante2017} is sent to the robot. For torque-controlled robot, $\tau$ is used as command. 

The problem should give enough constraints on the motion to specify a single solution, and not let the numerical solver arbitrate and possibly induce discontinuities between the solutions for two consecutive states. A common remedy is to use regularization tasks (typically a posture task, $\lambda = 0$, or $\tau = 0$). In practice, the problem will be over-constrained, meaning that some tasks are in conflict, and the design of the controller needs to specify how to arbitrate the conflicts during the resolution (see Section~\ref{sec:conflicts}).

\subsubsection{Pre-solving variants}
Problem \eqref{eq:instantaneous_problem} is written over all $3$ variables $\qddot$, $\tau$ and $\lambda$. This gives great flexibility to express new tasks/constraints and offers a clear vision of what we want to achieve.
However, the specific structure, of \eqref{eq:ocp_wb_dyn2} and \eqref{eq:contact_tasks} can be exploited to remove some variables and constraints, either to speed-up computation or for formulation-related reasons, giving rise to different families of controllers. This variable elimination can be regarded as partially pre-solving manually the problem before handling it to the solver.

The simple form of $S$ (e.g., $\begin{bmatrix}0 & I \end{bmatrix}$ for legged robot) allows to directly express $\tau$ as a linear function of $\qddot$ and $\lambda$ and eliminate it from the problem. Only the non-actuated rows of \eqref{eq:ocp_wb_dyn2} are kept. This is the approach in\cite{Herzog2016,Vaillant2016} and leads to an easy decrease of computation time.
Alternatively, \eqref{eq:ocp_wb_dyn2} can be used to eliminate $\qddot$ (and removed), leveraging the positive definite characteristic of $M$ for a cheap resolution based on the Cholesky decomposition\cite{Mansard2012, Wensing2013g}.

When $J_c = J$\footnote{this depends on the representation of the contact forces}, \eqref{eq:contact_tasks} can be used, in conjunction with \eqref{eq:ocp_wb_dyn2}, to remove $\lambda$. This projection on the nullspace of the contact constraints is used in~\cite{Park2006, Sentis2010} to extend the operational space control to underactuated robots \cite{righetti2011inverse}. 
More generally, one can use \eqref{eq:ocp_wb_dyn2} and \eqref{eq:contact_tasks} to solve for $\qddot$ and $\lambda$, keeping $\tau$ as the only variable.

Some variable eliminations are rooted in the history of families of controllers. But when it comes to performance, some points have to be kept in mind. Solvers are internally performing operations akin to variable eliminations. Therefore performing one manually is only useful when some properties of the matrices (e.g., sparsity or positive-definiteness) can be leveraged, that the solver cannot know. Solvers, on the other hand, can usually better handle numerical issues. The number of variables can be a good indicator of the computational cost, but the structure of the problem is also important~\cite{Dimitrov2014}. Removing a variable can destroy the structure, preventing a solver to exploit it. The simplest example is when removing a variable transforms bound constraints into general constraints. Overall, the optimal pre-solving variant is highly coupled with structure exploiting capabilities of the numerical solver used.


\subsubsection{Arbitrating conflict}
\label{sec:conflicts}
For most task sets considered in practice, tasks requirements will conflict, meaning \eqref{eq:instantaneous_problem} has no solution. In this case, one can associate a violation measure to each constraint, based on the $L2$ norm, and specify how important it is to minimize each violation. Two main approaches exist that can be combined: weighing each violation (also known as \emph{soft priorities}) or defining a hierarchy between them (\emph{strict priorities}). 

Numerous works use the so-called QP approach\footnote{technically constrained linear least-squares}: all inequality constraints and some equality constraints in \eqref{eq:instantaneous_problem} are kept as such in the QP, effectively being assigned the top priority. The weighted sum of the square norms of all other constraint violations forms the objective of the QP. For example, let us consider the problem of tracking a LIP-based reference trajectory from a simple model, and while trying to do so by using minimal joint torques within bounds. In this case, a QP could be formulated as:
\begin{align*}
    \minimize_{\qddot, \tau, \lambda}~&  w_1\| J_{G} \qddot + \dot{J}_G \nu - (\ddot{p}_G^d +  \ddot{e}^d_{G}) \|^2 + w_2 \|\tau\|^2 \\
    {\rm subject~to}~& M \qddot + C \qdot + \tau_g = S^T \tau + J_c^T \lambda ~~\textrm{(dynamics)}  \\
                     & J_c \qddot = a_c ~~\textrm{(fixed~contacts)} \\
                     & C \lambda \le 0 ~~\textrm{(friction~cones)}\\
                     &\ldots\textrm{(other~top-priority~constraints)}\ldots
\end{align*}

The use of QP formulations for reactive control is now nearly universal, but follows a long historical development. After an early work motivated by the inclusion of unilateral contact forces~\cite{Y.Fujimoto1998}, the QP-based inverse dynamics approach really started with~\cite{Abe2007,Collette2007} and closely related convex formulations \cite{Park2007}, continuing with~\cite{Bouyarmane2011,Lee2012, Koolen2013,Feng2015,Vaillant2016} among many others. An advantage of this approach is that it relies on off-the-shelf, mature solvers, largely available and often free~(e.g.,~\cite{ferreau2014qpoases,stellato2020osqp}) 
Its main limitations are the inability to handle inequalities at lower priority levels and that the tuning of weights becomes difficult when many tasks exist.

One of the benefits of the QP formalism is its flexibility to incorporate other control paradigms into reactive whole-body control. For example \cite{Kuindersma2016, chignoli2020variational} embed optimal control for simple-model tasks via a cost function term designed to descend a task-space value function. In this regard, the instantaneous nature of the QP is mitigated by the long-term performance considerations embedded in the value function. Other work has explored the incorporation of passivity-based control approaches that tweak the above problem in subtle ways that improve robustness to unmodeled effects \cite{Englsberger2020}. To improve robustness through other means, recent QP strategies have explored methods for controls to be invariant to velocity jumps from  impact events \cite{Yang2021}, which are hard to detect in practice. Alternative strategies that add the consideration of soft contact can also be incorporated into whole-body QPs \cite{fahmi2020stance,Flayols2021}.

In contrast to using a weighting approach for tasks, a fully hierarchical approach lets the designer specify an explicit priority level for each task. Tasks at the same level can be combined with weights, and top priority constraints need not be feasible. The resulting problem is called \emph{lexicographic least-squares} or \emph{hierarchical QP} (HQP) and used in numerous works\cite{DeLasa2010, Saab2013, Herzog2016}. The cascade of QP of Kanoun\cite{Kanoun2009} was the first solver able to tackle inequality constraints at any priority level, followed by variants improving its computation time\cite{DeLasa2010,Herzog2016}. 
 A few dedicated solvers\cite{Escande2014,Dimitrov2015} have been introduced, solving the problem efficiently in one pass.
The hierarchical formulation is a strict superset of the QP-based formulation, as well as the limit case for when the ratio between task weights goes to infinity~\cite{VanLoan1985}. It is shown in \cite{Dimitrov2015}, contrary to the common belief, a HQP can be solved faster using a dedicated solver, both in theory and practice. However, the computational cost of the cascade solvers, and the complexity of handling singularities \cite{Pfeiffer2018} have limited the use of the hierarchical approach. 




\subsection{Summary}

This section has introduced common methods for reactive control that coordinate the selection of joint torques and contact forces at the current instant. These methods complement predictive control (i.e., over a horizon) as provided by methods in the previous section. By focusing on an instantaneous control problem, the formulation inherits desirable structure (e.g., nonlinear constraints become linear ones). This enables solutions at rapid rates which are amenable to computation using off-the-shelf QP solvers. There are many different variants to such whole-body problems that consider soft priorities vs. hard priorities between tasks, or that strategically pre-solve for some decision variables to accelerate the QP solution. While these methods have represented a natural outgrowth of operational-space control to legged robotics applications, the shift from solving least-square problems to solving QPs has enabled many other control paradigms to be considered with WBC, including CLFs, CBFs, passivity-based strategies, and others guided by value functions. Overall, while there remain opportunities to accelerate these solution methods and improve their robustness, the current state of the art represents a reliable technology for current and future applications. 


\section{ Outlook and Prospects}
\label{sec:conclusion}


We conclude this survey by reflecting on some of the main trends that were identified through the literature review, and also some of the notable gaps that remain to be addressed. While many of these observations represent desired further improvements to model-based approaches, we close with a few final remarks on the outlook for model-based approaches within the context of the parallel growing inertia of data-driven approaches (e.g., RL) that have also had impressive recent demonstrations (e.g., \cite{lee2020learning,miki2022learning}). 

\subsection{Main Trends}

\paragraph*{Main trends for optimization with contacts} The past five years have seen an increase in contact implicit approaches that avoid specifying gait/contact sequences a-priori. These methods have considered either complaint contact or MIP/LCP strategies. Solution speed remains a challenge for these methods, as does non-convexity of the problem. Strategies that combine planning and/or learning with contact implicit methods appear as a necessary next step. In the near term, fixing contact sequences/timing remains the most viable option for practical deployment in applications that require online optimization (e.g., in model-predictive control). In the end, while much more recent work has concentrated on addressing rigid contacts, other approaches end up being similar since rigid complementarity constraints are often relaxed and smoothed, leading to similar operations as with compliant models.

\paragraph*{Main trends in the dynamic models adopted for model-based optimization} The past five years have observed an increase in model complexity used for MPC (with several examples using the full model). This advance has occurred from increased understanding of gradient-based numerical methods and the availability of fast open-source software libraries to compute robot's dynamics and its derivatives~\cite{Carpentier2019}. With Moore's law ending, subsequent advances may center on parallelization. However, even with increases in computational speed, the nonconvexity of problems is an unavoidable challenge with complex models. This observation suggests that simple models will likely remain pertinent for the near future. The review points that combination with learning strategies that include exploration (as in conventional RL) presents promise for optimization-based strategies to avoid poor local minima when adopting complex models.

\paragraph*{Main trends for transcription} The past five years have included emphasis on DDP methods as well as extensions of collocation schemes to contact-constrained/implicit settings. Adoption of DDP has been motivated in part by its demonstrated scalability to high-DoF systems. However, recent work has also shown specialized Riccati-like solvers \cite{frison2020hpipm}, which have the promise to bring many of the advantages of shooting via DDP to other transcription approaches. While the numerical methods are improving, convergence to infeasible solutions is a challenge for complex systems, and this holds with any of the transcription methods. The review points that addressing these challenges is an important open point for future work. Further, simultaneous methods (collocation and multiple shooting) will present additional opportunities for algorithm-level parallelism to take advantage of the shift in the how computational resources are expected to grow in the coming decades.

\subsection{Overview of Open Problems}


\paragraph*{Formal Analysis} Despite a great deal of recent progress, there is a broad lack of formal analysis regarding the properties and operation of existing optimization-based robot control paradigms. Existing theoretical results regarding the stability and recursive feasibility of nonlinear MPC are mostly focused on the simplified problem of regulation to an equilibrium or tracking a feasible reference state trajectory \cite{rawlings2017model}, which leaves a theoretical gap for the classes of problems considered in the locomotion community. Another barrier to meaningful analysis is a relative lack of work addressing considerations of robustness to bounded or stochastic uncertainty (e.g., \cite{Morimoto2003, Brasseur2015, gazar2021stochastic}). Such a treatment would, for example, be critical to understand how our control frameworks interact with state estimation. While there are clear academic open questions, it remains an open question of its own as to how much additional performance such analysis would unlock.


\paragraph*{Reducing the Expertise Required} Beyond the technical expertise required for model-based control, there remains a great deal of domain-specific expertise required for crafting cost functions and constraints. It remains a challenge to remove the human as an outer loop in the tuning of problem formulations. A part of the issue is that it is often not apparent how to write down cost functions for even simple tasks. For example, the problem of opening a door has binary success, and such cost structures are largely incompatible with current tools. Whether through automated methods, or through new tools that enable a broader set of cost designs, the best path for reducing required domain-specific expertise remains open.



\paragraph*{Optimal Control Versus or With RL?}

With the recent progress in RL for legged locomotion (e.g., \cite{lee2020learning,miki2022learning}) much debate has surrounded whether real-time optimization or reinforcement learned policies are the path forward.
RL methods have some clear benefits, the main one probably being their generality. Indeed, being gradient free, they have no trouble with non-smooth contact-implicit problems, which are instead  hard to solve with gradient-based optimization. An interesting explanation of why that is the case has been recently provided in \cite{Lidec2022}, drawing a connection between \emph{Policy Gradient} and the stochastic optimization technique \emph{Randomized Smoothing}. Although our community has grown attached to gradient-based optimization, in order to overcome the challenges discussed in this survey it may be worth exploring gradient-free approaches, with RL being among the most promising.
Another positive feature of RL methods (especially those based on Dynamic Programming~\cite{Lillicrap2015}) is their exploration ability, which is fundamental to solve non-convex problems in cases where TO struggles.

Still, there remain exciting opportunities for how model-based optimization can help accelerate learning. 
Strategies such as the popular guided policy search method \cite{levine2013guided} and relatives (e.g., \cite{zhao2022adversarially}) can leverage TO to cut down on needed exploration and guide learning. Another important opportunity for future work is to use the efficient constraint-handling nature of TO to help make learning \emph{safe} \cite{zanon2020safe}.  Overall, since RL and OCP methods try to solve similar problems (especially in offline simulation settings where models are heavily used), the history of their synergy is likely still quite early. 


\subsection{Closing Remarks}

Reflecting back on the trajectory of the field, we see a great deal of commonality on the transformation that is taking place now as occurred in the mid 2010s. In 2007, convex optimization strategies for reactive control were beginning to appear, with no shortage of concern regarding both (a) their online computation requirements or (b) the abandonment of previously available closed-form control solutions. By 2015, these methods were common place. 

From where we are at today, early demonstrations of whole-body predictive control on hardware have greatly benefited from the structural understanding of our physics-based models from previous reactive control developments, while also adopting reactive methods as low-level controllers. Yet, legitimate concerns remain regarding (a) online computational requirements and (b) the abandonment of convexity properties that would give solution assurances. 

While we hope these methods will become technologies in the near future, we also look forward to their role in shaping the next-generation of learning-based controllers, where shared optimal control fundamentals will allow some of the insights in formulation and methodologies to inform and accelerate RL solutions. Concerns regarding (a) offline computation requirements and (b) a lack of generalization guarantees with RL seem to both be well addressed by structure exploiting methods from model-based optimization, and we hope this survey will play a small role in helping those solutions come to light. This evolutionary progress shows a steady maturation of the field toward applications in logistics, agriculture, construction, and yet others. With external commercial funding at unprecedented levels, we look forward to the next decade of progress from legged robots, which promises to be greater than the previous one.

\bibliographystyle{IEEEtran}
\bibliography{main.bbl}

\end{document}